\documentclass[lettersize,journal]{IEEEtran}
\usepackage{amsmath,amsfonts}
\usepackage{algorithmic}
\usepackage{array}
\usepackage[caption=false,font=normalsize,labelfont=sf,textfont=sf]{subfig}
\usepackage{textcomp}
\usepackage{stfloats}
\usepackage{url}
\usepackage{verbatim}
\usepackage{graphicx}
\usepackage{balance}
\usepackage{mathtools}
\usepackage{gensymb}
\usepackage[export]{adjustbox}
\usepackage{color}
\usepackage{graphics} 
\usepackage{epsfig}
\usepackage{times}
\usepackage{amssymb} 
\usepackage{leftidx}
\newtheorem{theorem}{Theorem}
\newtheorem{remark}{Remark}

\newtheorem{property}{Property}

\usepackage{siunitx}
\usepackage{cite}[noadjust]
\usepackage{nomencl}

\makenomenclature

\DeclareMathOperator{\diag}{diag}
\newcommand{\RomanNumeralCaps}[1]
    {\MakeUppercase{\romannumeral #1}}

\usepackage{algorithm}

\begin{document}

\title{
Modular Adaptive Aerial Manipulation under Unknown Dynamic Coupling Forces
}

\author{
Rishabh Dev Yadav$^{1}$, Swati Dantu$^{2}$, Wei Pan$^{1}$, Sihao Sun$^{3}$,  Spandan Roy$^{4}$, and Simone Baldi$^{5}$
\thanks{This work was supported in part by ``Aerial Manipulation'' under IHFC grand project (GP/2021/DA/032), in part by ``Capacity building for
human resource development in Unmanned Aircraft System (Drone and related Technology)'', MeiTY, India, in part by the Natural Science Foundation of China grants 62233004 and 62073074, and in part by Jiangsu Provincial Scientific Research Center of Applied Mathematics grant  BK20233002. Wei Pan acknowledges travel support from the European Union’s Horizon 2020 research and innovation programme under Grant Agreement No 951847. The first two authors contributed equally. \textit{(Corresponding authors: S. Baldi and S. Roy)}  }
\thanks{$^{1}$ The authors are with Department of Computer Science, The University of Manchester, UK {(email: rishabh.yadav@postgrad.manchester.ac.uk, wei.pan@manchester.ac.uk).}   }
\thanks{$^{2}$ S. Dantu is with Multi-robot systems group, Faculty of Electrical Engineering, Czech Technical University in Prague, Czech Republic.  {(email: dantuswa@fel.cvut.cz).}   }
\thanks{$^{3}$ S. Sun is with Department of Cognitive Robotics, Delft Univesrity of Technology, Netherlands {(email: s.sun-2@tudelft.nl).}   }
\thanks{$^{4}$ S. Roy is with Robotics Research Center, International Institute of Information Technology Hyderabad, India {(email: spandan.roy@iiit.ac.in).}  }
\thanks{$^{5}$ S. Baldi is with the School of Mathematics, Southeast University, Nanjing,
China (e-mail: simonebaldi@seu.edu.cn).
}
}

\markboth{{accepted to IEEE/ASME Transactions of Mechatronics. DOI: 10.1109/TMECH.2024.3457806 }}%
{Shell \MakeLowercase{\textit{et al.}}: A Sample Article Using IEEEtran.cls for IEEE Journals}

\maketitle

\begin{abstract}
Successful aerial manipulation largely depends on how effectively a controller can tackle {\color{black}the coupling dynamic forces between the aerial vehicle and the manipulator.} However, this control problem has remained largely unsolved as the existing control approaches either require precise knowledge of the aerial vehicle/manipulator inertial couplings, or neglect the state-dependent uncertainties especially arising during the interaction phase. This work proposes an adaptive control solution to overcome this long standing control challenge without any a priori knowledge of the coupling dynamic terms. Additionally, in contrast to the existing adaptive control solutions, the proposed control framework is modular, that is, it allows independent tuning of the adaptive gains for the vehicle position sub-dynamics, the vehicle attitude sub-dynamics, and the manipulator sub-dynamics. Stability of the closed loop under the proposed scheme is derived analytically, and real-time experiments validate the effectiveness of the proposed scheme over the state-of-the-art approaches. 
\end{abstract}


\section{INTRODUCTION}
An Unmanned Aerial Manipulator (UAM) is a coupled system where a quadrotor (or multirotor) vehicle carries a manipulator: the presence of {\color{black} the} manipulator greatly improves the dexterity and flexibility of the quadrotor, making it capable to accomplish a wide range of tasks, from simple payload transportation to more complex tasks such as pick and place, contact-based inspection, grasping and assembling etc. \cite{fumagalli2014developing, fumagalli2016mechatronic, mimmo2020robust, loianno2018localization, zhong2019practical, tognon2019truly, suarez2020benchmarks, li2023pseudospectral}. 

{\color{black}The UAM control problem faces considerable challenges due to the uncertain effects arising from (i) nonlinear coupling {\color{black}forces/wrenches} between the quadrotor and the manipulator  \cite{orsag2017dexterous, cao2019faster} and (ii) dynamic changes in the center of mass and mass/inertia distribution when the manipulator arm swings and/or interacts with the environment (during grasping, pick-and-drop, etc.)  \cite{tilli2021low, ruggiero2018aerial, xiao2021prescribed}.} Such inertial couplings are inescapable yet difficult to model with sufficient precision due to their state dependency. In the following, we discuss the state-of-the-art controllers developed for UAM and their limitations which lead toward the motivation of this work.

\subsection{Related Works and Motivation}
As observed in the reviews \cite{ruggiero2018aerial, ollero2021past}, the control approaches neglecting the coupling dynamic forces (cf. \cite{jimenez2013control, ruggiero2015multilayer }) or the approaches requiring precise knowledge of the system model (cf. \cite{suarez2018physical, kim2018cooperative,zhang2019robust, lee2020aerial, bicego2020nonlinear}) often fail to meet the required performance. Therefore, although in limited numbers, various approaches have been recently developed to tackle model uncertainties and external disturbances (e.g. payload wrench, wind) either by robust control methods such as linear estimator \cite{zhang2019robust}, extended high-gain observer \cite{kim2017robust}, disturbance observer \cite{chen2020robust}, RISE-based method \cite{lee2022rise}, or by adaptive control methods such as adaptive disturbance observer \cite{liang2021low, liang2024observer}, adaptive backstepping \cite{li2024finite}, adaptive sliding mode observer \cite{chen2022adaptive}, adaptive sliding mode control \cite{kim2016vision, liang2022adaptive}. {\color{black}However, robust control methods require nominal knowledge of the system model and on bounds of the uncertainties, while most adaptive control methods require the uncertainties or their time-derivatives to be bounded a priori (cf. \cite{liang2021low, liang2024observer, chen2022adaptive, kim2016vision, li2024finite}), which cannot capture state-dependent uncertainties. It is worth mentioning that even when the a priori boundedness requirement is removed as in \cite{liang2022adaptive}, precise knowledge of the inertial couplings is required. Unfortunately, dynamic changes in the center of mass and mass/inertia distribution are very difficult, if at all possible, to model and the same holds for the dynamic coupling forces between the quadrotor and the manipulator (cf. \cite[Ch. 5.3]{orsag2018aerial} and detailed discussion in Remark \ref{rem_dist} later). Hence, the issue of unknown state-dependent inertial forces in aerial manipulation remains largely unsolved in the literature. }

Another issue {\color{black}in adaptive control methods} is that the adaptive gains are common to all subsystems (cf. \cite{liang2021low, chen2022adaptive, kim2016vision, liang2022adaptive}), i.e., to the manipulator sub-dynamics, the position sub-dynamics and the attitude sub-dynamics of the quadrotor. As a result, it becomes impossible to tune the gains to improve the performance of one subsystem without affecting the performance of all the other subsystems. The flexibility to tune control performance at the subsystem level is lost. 

\subsection{Contributions of This Study}
Summarizing the above discussions, the state-of-the-art adaptive control designs for UAM face two bottlenecks: (i) coping with unknown state-dependent dynamics terms including inertial coupling terms, and (ii) designing noninterlacing adaptive laws for the UAM subsystems.  
Toward this direction, a modular adaptive control framework is presented, which brings out the following novelties:
\begin{itemize}
\item We propose an adaptive control scheme that does not require a priori knowledge (nominal or upper bounds) of state-dependent uncertainties (unlike \cite{kim2017robust, zhang2019robust, chen2020robust,lee2022rise, liang2021low, chen2022adaptive, kim2016vision }) or of coupling inertial dynamic terms (unlike \cite{jimenez2013control, ruggiero2015multilayer, liang2022adaptive}).
\item We propose novel modular adaptive laws that allow tuning the adaptive gains of the quadrotor position sub-dynamics, quadrotor attitude sub-dynamics and manipulator sub-dynamics independent to each other (unlike \cite{liang2021low, chen2022adaptive, kim2016vision, liang2022adaptive}).
\end{itemize}
{\color{black} Associated with} the control design is a novel stability analysis that, in contrast to the state-of-the-art, captures a new way to model the uncertainties and novel auxiliary gains that take into account the couplings between the UAM subsystems. Comparative real-time experimental results demonstrate the effectiveness of the proposed scheme over the existing solutions.

The rest of the paper is organised as follows: Sect. \RomanNumeralCaps{2} describes the UAM dynamics and the control problem; Sect. \RomanNumeralCaps{3}  details the proposed control framework and its stability analysis respectively; Sect. \RomanNumeralCaps{4} provides comparative experimental results while Sect. \RomanNumeralCaps{5} provides concluding remarks.

\section{System Dynamics and Problem Formulation}
\begin{figure}[]
\begin{center}
    \includegraphics[scale=0.35]{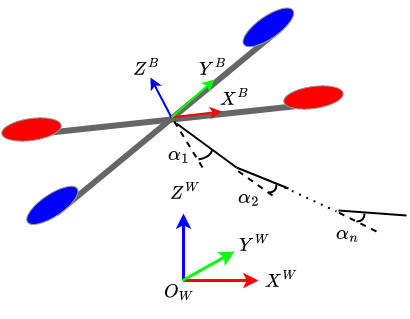}
    \caption{Schematic for a quadrotor-based UAM system with an $n$-link manipulator and the corresponding frames.}
    \label{robot_model}
\end{center}
\end{figure}

\begin{table}[!t]
\renewcommand{\arraystretch}{1.0}
\caption{{Nomenclature}}
\label{table_nomenclature}
\centering
{
{	\begin{tabular}{c c}
		\hline \\
	$[ X^B ~ Y^B ~Z^B ]$& Quadrotor body-fixed coordinate frame\\
	$[ X^W ~ Y^W ~ Z^W]$ & Earth-fixed coordinate frame \\
	$p=[x ~ y ~z]^{\top}$ & Quadrotor position in $[ X^W ~ Y^W ~ Z^W]$\\
	$q=[\phi ~ \theta ~ \psi]^{\top}$ & Quadrotor roll, pitch and yaw angles\\
	$\alpha=[\alpha_1, ~ \alpha_2, ~\cdot \cdot ~, \alpha_n]^{\top}$ & Manipulator joint angles\\
	$\boldsymbol M\in \mathbb{R}^{(6+n)\times (6+n)}$ & Mass matrix \\ 
	$\boldsymbol C\in \mathbb{R}^{(6+n)\times (6+n)}$ &  Coriolis matrix \\ 
	$g\in \mathbb{R}^{6+n}$&  Gravity vector \\
	$d\in \mathbb{R}^{6+n} $&  Bounded external disturbance \\
	$\tau_p, \tau_q \in \mathbb{R}^3$ & Generalized quadrotor control inputs \\
 $\tau_{\alpha} \in \mathbb{R}^n$ & Manipulator's joint control inputs \\
\hline 
\end{tabular}}}
\end{table}
The following notations are used in this paper:  $|| (\cdot)||$ and $\lambda_{\min}(\cdot)$ denote 2-norm and minimum eigenvalue of $(\cdot)$, respectively; $\boldsymbol I$ denotes identity matrix with appropriate dimension and $\diag\lbrace \cdot, \cdots, \cdot \rbrace$ denotes a diagonal matrix with diagonal elements $\lbrace \cdot, \cdots, \cdot \rbrace$.

Let us consider a quadrotor-based UAM system with an $n$ degrees-of-freedom (DoF) manipulator system as in Fig. \ref{robot_model} with symbols and system parameters given in Table \ref{table_nomenclature}, and having the Euler-Lagrangian dynamical model as \cite{arleo2013control}
\begin{equation} \label{EL_dynamics}
\boldsymbol M(\chi(t))\ddot{\chi}(t) +\boldsymbol C(\chi(t), \dot{\chi}(t))\dot{\chi}(t) + g(\chi(t)) + d(t) = \tau,    
\end{equation}
where $\chi = \begin{bmatrix} p^{\top} & q^{\top} & \alpha^{\top}\end{bmatrix}^{\top}, \tau = \begin{bmatrix}
    \tau_p^{\top} &    \tau_q^{\top} &    \tau_\alpha^{\top} \end{bmatrix}^{\top}\in \mathbb{R}^{6+n}$. 
The system dynamics terms can be decomposed as
\begin{subequations}\label{split_2}
\begin{align}
&\boldsymbol M = \begin{bmatrix}
    \boldsymbol M_{pp} & \boldsymbol M_{pq} & \boldsymbol M_{p\alpha} \\
    \boldsymbol M_{pq}^{\top} & \boldsymbol M_{qq} & \boldsymbol M_{q\alpha} \\
    \boldsymbol M_{p\alpha}^{\top} & \boldsymbol M_{q\alpha}^{\top} & \boldsymbol M_{\alpha \alpha}
    \end{bmatrix},~\begin{matrix}
\boldsymbol M_{pp}, \boldsymbol M_{qq}, \boldsymbol M_{pq} \in \mathbb{R}^{3\times3}\\ 
\boldsymbol M_{p\alpha}, \boldsymbol M_{q\alpha} \in \mathbb{R}^{3 \times n}\\ 
\boldsymbol M_{\alpha \alpha} \in \mathbb{R}^{n \times n}
\end{matrix}\\
&\boldsymbol C = \begin{bmatrix}
    \boldsymbol C_p \\
     \boldsymbol C_q \\
    \boldsymbol C_\alpha
    \end{bmatrix}, ~\begin{matrix} \boldsymbol C_p , \boldsymbol C_q \in \mathbb{R}^{3 \times (6+n)} \\ \boldsymbol C_\alpha \in \mathbb{R}^{n \times (6+n)}  \end{matrix} \label{new_dyn1}\\
&g = \begin{bmatrix}
    g_p \\
    g_q \\
    g_\alpha
    \end{bmatrix},
    d = \begin{bmatrix}
    d_p \\
    d_q \\
    d_\alpha
    \end{bmatrix}, ~\begin{matrix} g_p, g_q,d_p,d_q \in \mathbb{R}^{3}\\ g_\alpha, d_\alpha \in \mathbb{R}^{n}\end{matrix} \label{new_dyn2}
\end{align}
\end{subequations} 
Here $\tau_{\alpha} \triangleq
	\begin{bmatrix}
	\tau_{\alpha_1} & \tau_{\alpha_2}  & .. & \tau_{\alpha_n}
	\end{bmatrix}^{\top}$ is the control input for the manipulator; $\tau_q  \triangleq 
	\begin{bmatrix}
	u_2(t) & u_3(t) & u_4(t)
	\end{bmatrix}^{\top}$ is the control input for roll, pitch and yaw of the quadrotor; ${\tau_{p}} = {R^W_B U}$ is the generalized control input for quadrotor position in Earth-fixed frame, such that ${U}(t)\triangleq
	\begin{bmatrix}
	0 & 0 & u_1(t)
	\end{bmatrix}^{\top}\in \mathbb{R}^3$ is the force vector in body-fixed frame and $\boldsymbol {R}_B^W \in\mathbb{R}^{3\times3}$ is the $Z-Y-X$ Euler angle rotation matrix from the body-fixed coordinate frame to the Earth-fixed frame, given by
	\begin{align}
	\boldsymbol {R}_B^W =
	\begin{bmatrix}
	c_{\psi}c_{\theta} & c_{\psi}s_{\theta}s_{\phi} - s_{\psi}c_{\phi} & c_{\psi}s_{\theta}c_{\phi} + s_{\psi}s_{\phi} \\
	s_{\psi}c_{\theta} & s_{\psi}s_{\theta}s_{\phi} + c_{\psi}c_{\phi} & s_{\psi}s_{\theta}c_{\phi} - c_{\psi}s_{\phi} \\
	-s_{\theta} & s_{\phi}c_{\theta} & c_{\theta}c_{\phi}
	\end{bmatrix}, \label{rot_matrix}
	\end{align}
	where $c_{(\cdot)}$ and $s_{(\cdot)}$ denote $\cos{(\cdot)}$ {\color{black} and} $\sin{(\cdot)}$ respectively.

The following standard system properties hold from the Euler-Lagrange mechanics \cite{spong2008robot}:
\begin{property} \label{prop_1}
The matrix {$\boldsymbol M(\chi)$} is uniformly positive definite and $ \exists \underline{m},\overline{m} \in \mathbb{R}^{+}$ such that $0 < \underline{m}\boldsymbol I \leq \boldsymbol M(\chi) \leq \overline{m}\boldsymbol I$. 
\end{property}
 
\begin{property} \label{prop_2}
$\exists \bar{c}, \bar{g}, \bar{d}\in\mathbb{R}^{+}$ such that $||\boldsymbol C (\chi, \dot{\chi})|| \leq \bar{c}||\dot{\chi}||$, $||g (\chi)|| \leq \bar{g}$, $||d (t)|| \leq \bar{d}$. This implies, from (\ref{new_dyn1})-(\ref{new_dyn2}), $\exists \bar{c}_{p}, \bar{c}_{q}, \bar{c}_{\alpha}, \bar{g}_{p}, \bar{g}_{q}, \bar{g}_{\alpha}, \bar{d}_{p}, \bar{d}_{q}, \bar{d}_{\alpha} \in\mathbb{R}^{+}$ such that $||\boldsymbol C_{p} (\chi, \dot{\chi})|| \leq \bar{c}_{p}||\dot{\chi}||, ||\boldsymbol C_{q} (\chi, \dot{\chi})|| \leq \bar{c}_{q}||\dot{\chi}||, ||\boldsymbol C_{\alpha} (\chi, \dot{\chi})|| \leq \bar{c}_{\alpha}||\dot{\chi}||$, $||g_p (\chi)|| \leq \bar{g}_{p}, ||g_q (\chi)|| \leq \bar{g}_{q}, ||g_\alpha (\chi)|| \leq \bar{g}_{\alpha}$, $||d_p (t)|| \leq \bar{d}_{p}, ||d_q (t)|| \leq \bar{d}_{q}, ||d_\alpha (t)|| \leq \bar{d}_{\alpha}$.
\end{property}

In the following, we highlight the model uncertainties considered in this work:
\begin{remark}[Uncertainty]
Both the system dynamics terms $ \boldsymbol M, \boldsymbol C_{p}, \boldsymbol C_{q}, \boldsymbol C_{\alpha}, g_{p}, g_{q}, g_{\alpha}, d_{p}, d_{q}, d_{\alpha}$
and their bounds $\overline{m}, \underline{m}, \bar{c}_{p}, \bar{c}_{q}, \bar{c}_{\alpha}, \bar{g}_{p}, \bar{g}_{q}, \bar{g}_{\alpha}, \bar{d}_{p}, \bar{d}_{q}, \bar{d}_{\alpha}$ defined in Properties 1-2 are unknown for control design. Thus, we depart from state-of-the-art methods \cite{kim2017robust, zhang2019robust, chen2020robust,lee2022rise, liang2021low, chen2022adaptive, kim2016vision, liang2022adaptive} requiring a priori system knowledge.
\end{remark}

For the proposed modular control design, the UAM dynamics (\ref{EL_dynamics}) can be re-written using (\ref{split_2}) as
\begin{subequations} \label{dynamics}
\begin{align}
&\boldsymbol M_{pp}\ddot{p}\hspace{-.02cm}  + \hspace{-.02cm} \boldsymbol M_{pq}\ddot{q} + \hspace{-.02cm} \boldsymbol M_{p\alpha}\ddot{\alpha} + \hspace{-.02cm} \boldsymbol C_p\dot{\chi} \hspace{-.02cm} +\hspace{-.02cm} g_p \hspace{-.02cm}+ \hspace{-.02cm}d_p \hspace{-.02cm}=\hspace{-.01cm}  \tau_p,~ {\color{black}\tau_{p} \hspace{-.02cm}=\hspace{-.02cm} \boldsymbol{R}^W_B U} \label{pos}\\
&\boldsymbol M_{qq}\ddot{q}+\boldsymbol M_{pq}^{\top}\ddot{p}  + \boldsymbol M_{q\alpha}\ddot{\alpha} + \boldsymbol C_q\dot{\chi} + g_q + d_q = \tau_q \label{att} \\
&\boldsymbol M_{\alpha\alpha}\ddot{\alpha} + \boldsymbol M_{p\alpha}^{\top}\ddot{p} + \boldsymbol M_{q\alpha}^{\top}\ddot{q}  + \boldsymbol C_\alpha \dot{\chi} + g_\alpha + d_\alpha = \tau_\alpha \label{man} 
\end{align}
\end{subequations}
where (\ref{pos}), (\ref{att}) and (\ref{man}) represent the quadrotor position sub-dynamics, quadrotor attitude sub-dynamics and manipulator sub-dynamics along with their interactions, respectively.

{\color{black}\begin{remark}[Unknown inertial couplings] \label{rem_dist}
The inertial coupling terms $\boldsymbol M_{p\alpha}$ and $\boldsymbol M_{q\alpha}$ in (\ref{dynamics}) represent the interaction between the manipulator-quadrotor position sub-dynamics and the manipulator-quadrotor attitude sub-dynamics, respectively. 
These terms cause state-dependent forces (via acceleration terms $\ddot{p}, \ddot{q}, \ddot{\alpha}$) that, if neglected, may result in poor control performance or even instability. Unfortunately, these coupling terms are very difficult, if at all possible, to model in a uniform and precise way, as they depend on how the manipulator interacts with the environment (cf. \cite[Ch. 5.3]{orsag2018aerial}). This work departs from the existing literature (cf. \cite{kim2017robust, zhang2019robust, chen2020robust,lee2022rise, liang2021low, chen2022adaptive, kim2016vision, li2024finite, liang2024observer, liang2022adaptive}) as it does not neglect the inertial coupling terms, but avoids their knowledge by considering them as state-dependent uncertainties. 
\end{remark}}

The desired trajectories $\chi_d = \begin{bmatrix} p_d^{\top} & q_d^{\top} & \alpha_d^{\top}\end{bmatrix}^{\top}$ and their time-derivatives $\dot{\chi}_d, \ddot{\chi}_d$ are designed to be bounded. Furthermore, $\chi, \dot{\chi}, \ddot{\chi}$ are considered to be available for feedback. Let us define the tracking error as 
\begin{align}
  e \triangleq \chi(t) - \chi_d(t), {\color{black} ~\xi(t) \triangleq \begin{bmatrix}
	e^{\top}(t) & \dot{e}^{\top}(t)
	\end{bmatrix}^{\top} }. \label{err}
\end{align}

\textit{Control Problem:} Under Properties 1-2, to design an adaptive control framework for the UAM system (\ref{dynamics}) with modular adaptive laws {\color{black} to stabilize the tracking errors $(e, \dot{e})$ while tackling uncertainties described in Remark 1.}

The following section solves the control problem.

\section{Proposed Modular Adaptive Control Design and Analysis}
The proposed control framework consists of designing the quadrotor position control (Sect. III-A), the quadrotor attitude control (Sect. III-B) and the manipulator control (Sect. III-C) as per the dynamics (\ref{dynamics}). Note that, being the dynamics coupled, such design process is not decoupled: thus, Sect. III-D and the Appendix cover the overall stability analysis.
	
\subsection{Quadrotor Position Control}
For control design purpose, the quadrotor position sub-dynamics (\ref{dynamics}a) is rearranged as
\begin{equation} \label{dynamics_p}
\bar {\boldsymbol{M}}_{pp}\ddot{p} + E_p = \tau_p    
\end{equation}
where $\bar{\boldsymbol{M}}_{pp}$ is a user-defined constant positive definite matrix and $E_p \triangleq (\boldsymbol M_{pp} - \bar{\boldsymbol{M}}_{pp})\ddot{p} + \boldsymbol M_{pq}\ddot{q} + \boldsymbol M_{p\alpha}\ddot{\alpha} + \boldsymbol C_p\dot{\chi} + g_p + d_p$. The selection of $\bar{\boldsymbol{M}}_{pp}$ would be discussed later (cf. Remark \ref{mass}). Let us define the position tracking error as $e_p(t) \triangleq p(t) - p_d(t)$ and  
an error variable $r_p$ as
\begin{equation} \label{err_variable_p}
r_p \triangleq \boldsymbol B^{\top}_p \boldsymbol P_p \xi_p ,   
\end{equation}
where $\xi_p(t) \triangleq \begin{bmatrix}
	e^{\top}_p(t) & \dot{e}^{\top}_p(t)
	\end{bmatrix}^{\top}$, 
	$\boldsymbol B_p \triangleq \begin{bmatrix}
	\boldsymbol 0 & \boldsymbol I
	\end{bmatrix}^{\top}$; $\boldsymbol P_p > 0$ is the solution to the Lyapunov equation $\boldsymbol A^{\top}_p \boldsymbol P_p + \boldsymbol P_p \boldsymbol A_p = -\boldsymbol Q_p $ for some $\boldsymbol Q_p > \boldsymbol 0$ with $\boldsymbol A_p \triangleq \begin{bmatrix}
	\boldsymbol 0 & \boldsymbol I \\ -\boldsymbol{\lambda}_{p1} & -\boldsymbol{\lambda}_{p2} 
	\end{bmatrix}$. Here, $\boldsymbol{\lambda}_{p1}$ and $\boldsymbol{\lambda}_{p2}$ are two user-defined positive definite gain matrices and their positive definiteness guarantees that $\boldsymbol A_p$ is Hurwitz.

The quadrotor position control law is designed as
\begin{subequations}\label{ct1}
\begin{align}
\tau_p &= \bar{\boldsymbol{M}}_{pp}(-\boldsymbol \Lambda_p \xi_p - \Delta \tau_p + \ddot{p}_d),  \label{tau_p}\\
\Delta \tau_p &= \begin{cases}
    \rho_p \frac{r_p}{||r_p||}       & ~ \text{if } || r_p|| \geq \varpi_p\\
    \label{del_p}
    \rho_p \frac{r_p}{\varpi_p}       & ~ \text{if } || r_p|| < \varpi_p\\
    \end{cases},
\end{align}
\end{subequations}
where $\boldsymbol  \Lambda_p$ is a user-defined positive definite gain matrix and $\varpi_p > 0$ is a scalar used to avoid chattering; $\rho_p$ tackles the uncertainties, whose design will be discussed later.
Substituting (\ref{tau_p}) into (\ref{dynamics_p}) yields
\begin{equation}
\ddot{e}_p = -\boldsymbol  \Lambda_p \xi_p - \Delta \tau_p + \sigma_p, \label{err1}   
\end{equation}
where $\sigma_p \triangleq - \bar{\boldsymbol{M}}_{pp}^{-1} E_p$ is defined as the overall uncertainty. Using system Properties 1-2 one can verify
 \begin{align}
||\sigma_p || &\leq ||\bar{\boldsymbol{M}}_{pp}^{-1}|| (||(\boldsymbol M_{pp} - \bar{\boldsymbol{M}}_{pp} )|| ||\ddot{p}|| + || \boldsymbol M_{pq}|| ||\ddot{q}|| \nonumber \\ &+ || \boldsymbol M_{p\alpha}|| ||\ddot{\alpha}|| + ||\boldsymbol C_p|| ||\dot{\chi}|| + ||{g}_p|| + ||{d}_p|| ). \label{cp}
 \end{align}
Using Property 2 and  the inequalities $||\ddot{\chi}|| \geq ||\ddot{p}||$, $||\ddot{\chi}|| \geq ||\ddot{q}||$ and $||\ddot{\chi}|| \geq ||\ddot{\alpha}||$, $||\xi|| \geq ||\dot{e}||$, $||\xi|| \geq ||{e}||$ and substituting $\dot{\chi} = \dot{e} + \dot{\chi}_d$ into (\ref{cp}) yields
\begin{align} 
||\sigma_p|| &\leq K_{p0}^*  +K_{p1}^*||\xi||+ K_{p2}^*||\xi||^2 + K_{p3}^*||\ddot{\chi}||\label{up_bound_p} 
\end{align}
\begin{align*}
\text{where}~ K_{p0}^* &= ||\bar{\boldsymbol{M}}_{pp}^{-1}||(\bar{g}_{p} + \bar{d}_{p} + \bar{c}_{p}||\dot{\chi}_d||^2),\\
K_{p1}^* &= 2||\bar{\boldsymbol{M}}_{pp}^{-1}||\bar{c}_{p}||\dot{\chi}_d||,~
K_{p2}^* = ||\bar{\boldsymbol{M}}_{pp}^{-1}||\bar{c}_{p}, \\
K_{p3}^* &= ||\bar{\boldsymbol{M}}_{pp}^{-1}|| (||(\boldsymbol M_{pp} - \bar{\boldsymbol{M}}_{pp} )||+ ||\boldsymbol M_{pq}|| + ||\boldsymbol M_{p\alpha}||)
\end{align*}
are unknown scalars. Based on the upper bound structure in (\ref{up_bound_p}), the gain $\rho_p$ in (\ref{del_p}) is designed as
\begin{align}
\rho_{p} = \hat{K}_{p0} + \hat{K}_{p1}||\xi|| + \hat{K}_{p2}||\xi||^2 + \hat{K}_{p3}||\ddot{\chi}|| + \zeta_p,  \label{rho_p} 
\end{align}
where {\color{black}$\hat{K}_{pi}$ is the estimate of $K_{pi}^*$ for each $i=0,1,2,3$}, and $\zeta_p$ is an auxiliary stabilizing gain (cf. Remark \ref{rem_zeta}). The gains $\hat{K}_{pi}$ and $\zeta_p$ are adapted via the following laws:
\begin{subequations}\label{adaptive_law_p}
\begin{align}
&\dot{\hat{K}}_{p i} = ||r_p||||\xi||^i - \nu_{p i} \hat{K}_{p i},~~ i = 0,1,2 \\
&\dot{\hat{K}}_{p 3} =  ||r_p||||\ddot{\chi}|| - \nu_{p3} \hat{K}_{p3},\\
&\dot{\zeta}_p = \begin{cases}
    0 ~~~~~~~~ \text{if } || r_p|| \geq \varpi_p,  \\
    -\lbrace 1 + (\hat{K}_{p3}||\ddot{\chi}||+ \sum_{i=0}^{2}\hat{K}_{p i} ||\xi||^i )||r_p|| \rbrace \zeta_p + {\epsilon}_p, \\
  ~~~~~~~~~~ \text{if } || r_p|| < \varpi_p,
\end{cases}\\
& \qquad ~\hat{K}_{p i} (0) > 0,~\hat{K}_{p3} (0)>0,~\zeta_p (0) >0, \label{zeta1}
\end{align}
\end{subequations}
where $\nu_{p0}, \nu_{p1}, \nu_{p2}, \nu_{p3}, {\epsilon}_p \in\mathbb{R}^{+}$ are user-defined scalars.

\subsection{Quadrotor Attitude Control}
To achieve the attitude control, the tracking error in quadrotor attitude is defined as \cite{mellinger2011minimum,lee2010geometric}
\begin{align}
    e_q &= \frac{1}{2}{((\boldsymbol 
 R_d)^{\top} \boldsymbol R_B^W - (\boldsymbol R_B^W)^{\top} \boldsymbol R_d)}^{v}  \\
    \dot{e}_q & = \dot{q} - \boldsymbol R_d^{\top} \boldsymbol R_B^W \dot{q}_d
\end{align}
where $(.)^v$ is \textit{vee} map converting elements from $SO(3)$ to ${\mathbb{R}^3}$ and $\mathbf R_d$ is the rotation matrix as in (\ref{rot_matrix}) evaluated at ($\phi_d, \theta_d, \psi_d$).

The quadrotor attitude sub-dynamics (\ref{dynamics}b) is rearranged as
\begin{equation} \label{dynamics_q}
\bar{\boldsymbol M}_{qq}\ddot{q} + E_q = \tau_q , 
\end{equation}
where $\bar{\boldsymbol M}_{qq}$ is a user-defined constant positive definite matrix (cf. Remark \ref{mass}) and $E_q \triangleq (\boldsymbol M_{qq} - \bar{\boldsymbol M}_{qq})\ddot{q} + \boldsymbol M_{pq}^{\top}\ddot{p} + \boldsymbol M_{q\alpha}\ddot{\alpha} + \boldsymbol C_q\dot{\chi} + g_q + d_q$. 

Let us define an error variable $r_q$ as
\begin{equation} \label{err_variable_q}
r_q \triangleq \boldsymbol B^{\top} \boldsymbol P_q \xi_q  ,
\end{equation}
where $\xi_q(t) \triangleq \begin{bmatrix}
	e^{\top}_q(t) & \dot{e}^{\top}_q(t)
	\end{bmatrix}^{\top}$; $\boldsymbol P_q > \boldsymbol 0$ is the solution to the Lyapunov equation $\boldsymbol A^{\top}_q\boldsymbol P_q + \boldsymbol P_q \boldsymbol A_q = -\boldsymbol Q_q $ for some $\boldsymbol Q_q > \boldsymbol 0$ with $\boldsymbol A_q \triangleq \begin{bmatrix} \boldsymbol
	0 & \boldsymbol I \\ -\boldsymbol{\lambda}_{q1} & -\boldsymbol{\lambda}_{q2} 
	\end{bmatrix}$ and $\boldsymbol{\lambda}_{q1},\boldsymbol{\lambda}_{q2}$ being two user-defined positive definite gain matrices and their positive definiteness guarantees that $\boldsymbol A_q$ is Hurwitz.
 
The quadrotor attitude control law is proposed as
\begin{subequations}\label{ct2}
\begin{align} 
\tau_q &= \bar{\boldsymbol M}_{qq}(-\boldsymbol \Lambda_q \xi_q - \Delta \tau_q + \ddot{q}_d), \label{tau_q} \\
\Delta \tau_q &= \begin{cases}
    \rho_q \frac{r_q}{||r_q||}       & ~ \text{if } || r_q|| \geq \varpi_q\\
    \label{rho_q}
    \rho_q \frac{r_q}{\varpi_q}       & ~ \text{if } || r_q|| < \varpi_q\\
    \end{cases},
\end{align}
\end{subequations}
where $\boldsymbol \Lambda_q$ is a user-defined positive definite gain matrix and $\varpi_q > 0$ is a scalar used to avoid chattering; the design of gain $\rho_q$ will be discussed later.
Substituting (\ref{tau_q}) into (\ref{dynamics_q}) yields
\begin{equation}
\ddot{e}_q = -\boldsymbol \Lambda_q \xi_q - \Delta \tau_q + \sigma_q, \label{err2}   
\end{equation}
where $\sigma_q \triangleq - \bar{\boldsymbol M}_{qq}^{-1} E_q$ is defined as the uncertainty pertaining to attitude sub-dynamics. Following similar lines to derive the upper bound of $||\sigma_q||$ as in (\ref{up_bound_p}), one can derive
 \begin{align}
||\sigma_q || &\leq ||\bar{\boldsymbol M}_{qq}^{-1}|| (||(\boldsymbol M_{qq} - \bar{\boldsymbol M}_{qq} )|| ||\ddot{q}|| + || \boldsymbol M_{pq}^{\top}|| ||\ddot{p}|| \nonumber \\ &+ || \boldsymbol M_{q\alpha}|| ||\ddot{\alpha}|| + ||\boldsymbol {C}_q|| ||\dot{\chi}|| + ||{g}_q|| + ||{d}_q|| ) \nonumber\\
&\leq K_{q0}^* + K_{q1}^*||\xi|| + K_{q2}^*||\xi||^2 + K_{q3}^*||\ddot{\chi}||\label{up_bound_q} 
\end{align}
\begin{align*}
\text{where}~ K_{q0}^* &= ||\bar{\boldsymbol M}_{qq}^{-1}||(\bar{g}_{q} + \bar{d}_{q} + \bar{c}_{q}||\dot{\chi}_d||^2)\\
K_{q1}^* &= 2\bar{c}_{q}||\bar{\boldsymbol M}_{qq}^{-1}||||\dot{\chi}_d||,~
K_{q2}^* = ||\bar{\boldsymbol M}_{qq}^{-1}||\bar{c}_{q} \\
K_{q3}^* &=||\bar{\boldsymbol M}_{qq}^{-1}||( ||(\boldsymbol M_{qq} - \bar{\boldsymbol M}_{qq} )|| + ||\boldsymbol M_{pq}^{\top}|| + ||\boldsymbol M_{q\alpha}||)
\end{align*}
are unknown scalars. 
Based on the upper bound structure in (\ref{up_bound_q}), the gain $\rho_q$ in (\ref{rho_q}) is designed as
\begin{align} 
\rho_{q} &= \hat{K}_{q0} + \hat{K}_{q1}||\xi|| + \hat{K}_{q2}||\xi||^2 + \hat{K}_{q3}||\ddot{\chi}|| + \zeta_q, 
\end{align}
where $\zeta_q$ is a stabilizing gain and $\hat{K}_{qi}$ are the estimates of $K_{qi}^*$ for each $i=0,1,2,3$, adapted via the following laws:
\begin{subequations} \label{adaptive_law_q}
\begin{align} 
&\dot{\hat{K}}_{q i} = ||r_q||||\xi||^i - \nu_{q i} \hat{K}_{q i},~~ i = 0,1,2 \\
&\dot{\hat{K}}_{q 3} =  ||r_q||||\ddot{\chi}|| - \nu_{q3} \hat{K}_{q3},\\
&\dot{\zeta}_q = \begin{cases}
    0 ~~~~~~~~ \text{if } || r_q|| \geq \varpi_q,  \\
    -\lbrace 1 + (\hat{K}_{q3}||\ddot{\chi}||+ \sum_{i=0}^{2}\hat{K}_{q i} ||\xi||^i )||r_q|| \rbrace \zeta_q + {\epsilon}_q, \\
  ~~~~~~~~~~ \text{if } || r_q|| < \varpi_q,
\end{cases}\\
& \qquad ~\hat{K}_{q i} (0) > 0,~\hat{K}_{q3} (0)>0,~\zeta_q (0) >0, \label{zeta2}
\end{align}
\end{subequations}
where $\nu_{q0}, \nu_{q1}, \nu_{q2}, \nu_{q3},  {\epsilon}_q \in\mathbb{R}^{+}$ are user-defined scalars.

\subsection{Manipulator Control}
Similar to the previous subsections, the manipulator sub-dynamics (\ref{dynamics}c) is rearranged via introducing a user-defined constant positive definite matrix $\bar{\boldsymbol M}_{\alpha\alpha}$ (cf. Remark \ref{mass}) as
\begin{equation} \label{dynamics_alpha}
\bar{\boldsymbol M}_{\alpha\alpha}\ddot{\alpha} + E_\alpha = \tau_\alpha ,   
\end{equation}
where $E_\alpha \triangleq (\boldsymbol M_{\alpha\alpha} - \bar{\boldsymbol M}_{\alpha\alpha})\ddot{\alpha} + \boldsymbol M_{p\alpha}^{\top}\ddot{p} + \boldsymbol M_{q\alpha}^{\top}\ddot{q} + \boldsymbol C_\alpha\dot{\chi} + g_\alpha + d_\alpha $.
Taking $e_\alpha(t) \triangleq \alpha(t) - \alpha_d(t)$ and  $\xi_\alpha(t) \triangleq \begin{bmatrix}
	e^{\top}_\alpha(t) & {\color{black} \dot{e}^{\top}_\alpha(t)}
	\end{bmatrix}^{\top}$, an error variable $r_\alpha $ is defined as
\begin{equation} \label{err_variable_alpha}
r_\alpha \triangleq \boldsymbol B^{\top} \boldsymbol P_\alpha \xi_\alpha  ,  
\end{equation}
where $\boldsymbol P_\alpha > \boldsymbol 0$ is the solution to the Lyapunov equation $\boldsymbol A^{\top}_\alpha \boldsymbol P_\alpha + \boldsymbol P_\alpha \boldsymbol A_\alpha = - \boldsymbol Q_\alpha $ for some $\boldsymbol Q_\alpha > \boldsymbol 0$ with $\boldsymbol A_\alpha \triangleq \begin{bmatrix}
	\boldsymbol 0 & \boldsymbol I \\ -\boldsymbol{\lambda}_{\alpha1} & -\boldsymbol{\lambda}_{\alpha2} 
	\end{bmatrix}$. The matrices $\boldsymbol{\lambda}_{\alpha1}$ and $\boldsymbol{\lambda}_{\alpha2}$ are designed to be positive definite so that $\boldsymbol A_\alpha $ is Hurwitz.

The manipulator control law is designed as
\begin{subequations}\label{ct3}
\begin{align} 
\tau_\alpha &= \bar{\boldsymbol M}_{\alpha\alpha}(- \boldsymbol \Lambda_\alpha \xi_\alpha - \Delta \tau_\alpha + \ddot{\alpha}_d), \label{tau_alpha}\\
\Delta \tau_\alpha &= \begin{cases}
    \rho_\alpha \frac{r_\alpha}{||r_\alpha||}       & ~ \text{if } || r_\alpha|| \geq \varpi_\alpha\\
    \label{rho_alpha}
    \rho_\alpha \frac{r_\alpha}{\varpi_\alpha}       & ~ \text{if } || r_\alpha|| < \varpi_\alpha\\
    \end{cases},
\end{align}
\end{subequations}
where $\boldsymbol \Lambda_\alpha$ is a user-defined positive definite gain matrix and the scalar $\varpi_\alpha > 0$ is used to avoid chattering; the gain $\rho_\alpha$ is designed later.
Substituting (\ref{tau_alpha}) in (\ref{dynamics_alpha}) yields
\begin{equation}
\ddot{e}_\alpha = - \boldsymbol \Lambda_\alpha \xi_\alpha - \Delta \tau_\alpha + \sigma_\alpha    \label{err3}
\end{equation}
where $\sigma_\alpha \triangleq - \bar{\boldsymbol M}_{\alpha\alpha}^{-1} E_\alpha$ defines the overall uncertainty in manipulator sub-dynamics. Similar to the upper bounds of $||\sigma_p ||$ and $||\sigma_q ||$, one can verify
 \begin{align}
||\sigma_\alpha || &\leq ||\bar{\boldsymbol M}_{\alpha\alpha}^{-1}||( ||(\boldsymbol M_{\alpha\alpha} - \bar{\boldsymbol M}_{\alpha\alpha} )|| ||\ddot{\alpha}|| + || \boldsymbol M_{p\alpha}^{\top}|| ||\ddot{p}|| \nonumber \\ &+ || \boldsymbol M_{q\alpha}^{\top}|| ||\ddot{q}|| + ||\boldsymbol C_\alpha|| ||\dot{\chi}|| + ||{g}_\alpha|| + ||{d}_\alpha|| ) \nonumber\\
 &\leq K_{\alpha0}^* + K_{\alpha1}^*||\xi|| + K_{\alpha2}^*||\xi||^2 + K_{\alpha3}^*||\ddot{\chi}||,  \label{up_bound_alpha}
\end{align}
\begin{align*}
\text{with}~ K_{\alpha0}^* &= ||\bar{\boldsymbol M}_{\alpha\alpha}^{-1}||(\bar{g}_{\alpha} + \bar{d}_{\alpha} + \bar{c}_{\alpha}||\dot{\chi}_d||^2),\\
K_{\alpha1}^* &= 2\bar{c}_{\alpha} ||\bar{\boldsymbol M}_{\alpha\alpha}^{-1}||||\dot{\chi}_d||,~
K_{\alpha2}^* = ||\bar{\boldsymbol M}_{\alpha\alpha}^{-1}||\bar{c}_{\alpha}, \\
K_{\alpha3}^* &= ||\bar{\boldsymbol M}_{\alpha\alpha}^{-1}||( ||(\boldsymbol M_{\alpha\alpha} - \bar{\boldsymbol M}_{\alpha\alpha} )||+ ||\boldsymbol M_{p\alpha}^{\top}|| + ||\boldsymbol M_{q\alpha}^{\top}||)
\end{align*}
being unknown constants. Based on the upper bound structure in
(\ref{up_bound_alpha}), the gain $\rho_\alpha$ in (\ref{rho_alpha}) is designed as
\begin{align}
\rho_{\alpha} &= \hat{K}_{\alpha0} + \hat{K}_{\alpha1}||\xi|| + \hat{K}_{\alpha2}||\xi||^2 + \hat{K}_{\alpha3}||\ddot{\chi}|| + \zeta_\alpha ,
\end{align}
where $\zeta_\alpha$ is an auxiliary stabilizing gain and $\hat{K}_{\alpha i}$ are the estimates of ${K}_{\alpha i}^*$ for each $i=0,1,2,3$, adapted via the following laws:
\begin{subequations} \label{adaptive_law_alpha}
\begin{align}
&\dot{\hat{K}}_{\alpha i} = ||r_\alpha||||\xi||^i - \nu_{\alpha i} \hat{K}_{\alpha i},~~ i = 0,1,2 \\
&\dot{\hat{K}}_{\alpha 3} =  ||r_\alpha||||\ddot{\chi}|| - \nu_{\alpha3} \hat{K}_{\alpha3},\\
&\dot{\zeta}_\alpha = \begin{cases}
    0 ~~~~~~~~ \text{if } || r_\alpha|| \geq \varpi_\alpha,  \\
    -\lbrace 1 + (\hat{K}_{\alpha3}||\ddot{\chi}||+ \sum_{i=0}^{2}\hat{K}_{\alpha i} ||\xi||^i )||r_\alpha|| \rbrace \zeta_\alpha + {\epsilon}_\alpha,\\
  ~~~~~~~~~~ \text{if } || r_\alpha|| < \varpi_\alpha,
\end{cases}\\
& \qquad ~\hat{K}_{\alpha i} (0) > 0,~\hat{K}_{\alpha3} (0)>0,~\zeta_\alpha (0) >0, \label{zeta3}
\end{align}
\end{subequations}
where $\nu_{\alpha 0}, \nu_{\alpha 1}, \nu_{\alpha 2}, \nu_{\alpha 3}, {\epsilon}_\alpha \in\mathbb{R}^{+}$ are user-defined scalars. 

\begin{remark}[Choice of gains and trade-off]\label{mass}
It can be noted from (\ref{up_bound_p}), (\ref{up_bound_q}) and (\ref{up_bound_alpha}) that large values of the gains $\bar{\boldsymbol{M}}_{pp}, \bar{\boldsymbol M}_{qq}$ and $\bar{\boldsymbol M}_{\alpha \alpha}$ lead to lower values of  $K_{pi}^*$, $K_{qi}^*$ and $K_{\alpha i}^*$ $i=0,1,2,3$; consequently, the effects of unknown dynamics on the controller performance are reduced and faster adaptation is possible. Nevertheless, as a trade-off, large $\bar{\boldsymbol{M}}_{pp}, \bar{\boldsymbol M}_{qq}$ and $\bar{\boldsymbol M}_{\alpha \alpha}$ result in high control input demand (cf. (\ref{tau_p}), (\ref{tau_q}) and (\ref{tau_alpha})). {\color{black}Further, low values of $\nu_{pi}, \nu_{qi}, \nu_{\alpha i}$, $i=0,1,2,3$ in (\ref{adaptive_law_p}), (\ref{adaptive_law_q}) and (\ref{adaptive_law_alpha}) help in faster adaptation with higher adaptive gains, but at the cost of higher control input demand.  Therefore, the choice of these gains must consider application requirements and control input demand.}
\end{remark}

\subsection{Overall Control Structure and Stability Analysis}
The design steps of the proposed modular adaptive controller are summarized in Algorithm 1.
\begin{algorithm}[!h]
 \caption{ {Design steps of the proposed controller}}
 \vspace{0.2cm}
 
\textbf{Step 1 (Define the error variables):} Collect the error variables $(e_p, \dot{e}_p, \xi_p)$, $(e_q, \dot{e}_q, \xi_q)$ and $(e_\alpha, \dot{e}_\alpha, \xi_\alpha)$. Then, based on the selections of $\boldsymbol Q_p, \boldsymbol Q_q, \boldsymbol Q_\alpha$, find the solutions $\boldsymbol P_p, \boldsymbol P_q, \boldsymbol P_\alpha$ from the Lyapunov equations to calculate the error variables $r_p, r_q, r_\alpha$ as in \eqref{err_variable_p}, \eqref{err_variable_q} and \eqref{err_variable_alpha}.\\

\textbf{Step 2 (Evaluate adaptive gains):} 
After defining the appropriate gains $(\nu_{pi}, {\epsilon}_{p})$, $(\nu_{qi}, {\epsilon}_{q})$ and $(\nu_{\alpha i}, {\epsilon}_{\alpha})$, $i = 0, 1, 2, 3$, evaluate the adaptive gains $(\hat{K}_{pi}, \zeta_p)$, $(\hat{K}_{qi}, \zeta_q)$ and $(\hat{K}_{\alpha i}, \zeta_\alpha)$ from \eqref{adaptive_law_p}, \eqref{adaptive_law_q} and \eqref{adaptive_law_alpha}.\\

\textbf{Step 3 (Design control gains):} Design the positive definite gain matrices $(\bar{\boldsymbol M}_{pp}, \boldsymbol \Lambda_p)$, $(\bar{\boldsymbol M}_{qq}, \boldsymbol \Lambda_q)$ and $(\bar{\boldsymbol M}_{\alpha \alpha}, \boldsymbol \Lambda_\alpha)$. \\

\textbf{Step 4 (Design the controller):} Finally, using the results from Steps 1-3, evaluate the control inputs $\tau_p$, $\tau_q$ and $\tau_\alpha$ via \eqref{tau_p}, \eqref{tau_q} and \eqref{tau_alpha} respectively.

\vspace{0.2cm}
 \end{algorithm}
 
 \begin{theorem}
 Under the Properties \ref{prop_1}-\ref{prop_2}, the
closed-loop trajectories of (\ref{pos})-(\ref{man}) employing the control
laws (\ref{ct1}), (\ref{ct2}) and (\ref{ct3}) along with their respective adaptive laws (\ref{adaptive_law_p}), (\ref{adaptive_law_q}) and (\ref{adaptive_law_alpha}) are Uniformly Ultimately Bounded (UUB).
 \end{theorem}
\textit{Proof:}
 See Appendix.
 
\textit{Modularity in Adaptive Laws:}
The adaptive laws (\ref{adaptive_law_p}), (\ref{adaptive_law_q}) and \eqref{adaptive_law_alpha} reveal that the selection of the adaptive design parameters for one sub-dynamics is independent of the other ones via Step 1 (choice of $r_p, r_q, r_\alpha$ via different $\boldsymbol P_p, \boldsymbol P_q, \boldsymbol P_\alpha$) and Step 3 in Algorithm 1. Such a modularity allows tuning of the adaptive gain evaluations as per practical requirement as opposed to the use of common gains as in \cite{liang2021low, chen2022adaptive, kim2016vision, liang2022adaptive}. 
 
 \begin{figure*}
\includegraphics[width=\textwidth]{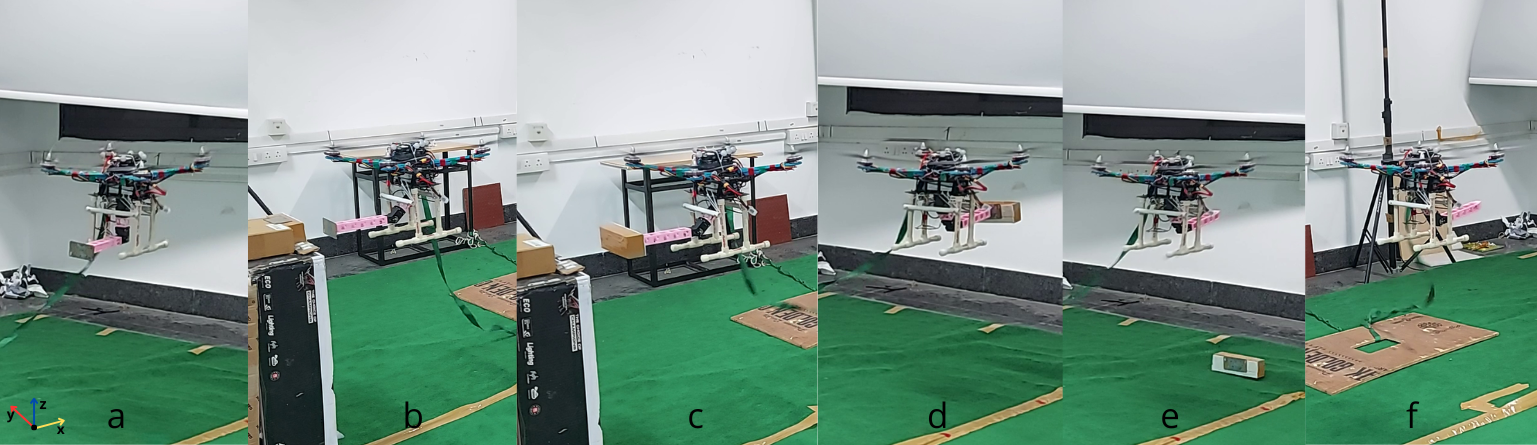}
    \centering
    \caption{Sequence of operations of the quadrotor during the experiment with the proposed controller: (a) takeoff from the ground; (b) follows the trajectory and expand arm to pick the payload; (c) picking the payload and stabilizing itself; (d) moving to the drop location while orienting the arm to the opposite direction; (e) dropping the payload and stabilizing itself (f) reaches to origin. The quadrotor is tied to the floor using a rope for safety reasons.}
    \label{fig:exp_snap}
\end{figure*}
 
\section{Experimental Results and Analysis}
For experimental purpose, a UAM setup is created using an S-500 quadrotor system (with brushless T-Motors 5010 {\color{black}750 KV}) and a 2R serial-link manipulator system (with Dynamixel XM430-W210-T motors). The manipulator carries an electromagnetic gripper for payload pick-and-place operation. The overall setup weighs 2.2 kg (approx). A U2D2 Power Hub Board is used to power the manipulator and the gripper. Raspberry Pi-4 is used as a processing unit which uses a U2D2 communication converter. {\color{black} Sensor data from OptiTrack motion capture system (at 120 fps) and IMU were used to measure the necessary pose (position, attitude), velocity and acceleration feedback of the drone. For the manipulator, the joint angular position and velocity are measured by the Dynamixel motors, while the accelerations are computed numerically.}

\subsection{Experimental Scenario}
To verify the performance of the proposed controller, an experimental scenario was created where the UAM is tasked to pick-up a payload (approx. $0.2$ kg) from a location and drop it to another designated place. The payload position was fixed and was made known to the quadrotor using the motion capture markers. The experimental scenario consists of the following sequences:
\begin{itemize}
    \item The quadrotor takes off from its origin (world coordinate $x = 0, y = 0$) to achieve a desired height $z_d=1$ m with the initial manipulator link angles $\alpha_1 (0) = \ang{0}$ and $\alpha_2(0) = \ang{90}$. 
    \item Quadrotor starts moving toward the payload (world coordinate $x = -1,y = 0$). During this time, the manipulator links start rotating to make desired angles $\alpha_1^d = \alpha_2^d = \ang{45}$ to pick the payload (at $2 \leq t \leq 30$ sec).
    \item After picking the payload at $t = 35$ sec (approx.) by activating the electromagnet, the quadrotor starts moving toward the payload drop point (world coordinate $x = 1,y = 0$). While moving toward the drop location, the manipulator links reorient to the opposite direction (at $35 < t \leq 65$ sec) to achieve $\alpha_1^d = \alpha_2^d = \ang{-45}$.
    \item After releasing the payload at $t = 70$ sec, the UAM returns to its origin ($x = 0, y = 0$).
\end{itemize}
Moving the manipulator arm while the quadrotor is in motion saves operational time since the arm can adjust to proper orientation to pick and drop the payload without putting the quadrotor into hovering condition. Nevertheless, this poses a steep control challenge as the center-of-mass of the system varies rapidly, and the inertial coupling dynamic forces become more prominent as $\ddot{p} \neq 0, \ddot{q} \neq 0 $ unlike the hovering condition. The electromagnet is programmed a priori to activate at $t=35$ sec and to deactivate at $t=70$ sec for picking and releasing the payload, respectively. 

To properly highlight the benefits of the proposed adaptive control design, the performance of the proposed controller is compared with the non-modular control techniques such as adaptive sliding mode controller (ASMC-1) \cite{kim2016vision} and (ASMC-2) \cite{liang2022adaptive}.  While ASMC-1 ignores state-dependent uncertainty, ASMC-2 can partially handle them as it requires precise knowledge of the mass matrix. A diagonal matrix was created for ASMC-1 and ASMC-2 based on the knowledge of mass of drone, mass of payload, arm link lengths etc. following the conventional system dynamics as in \cite{arleo2013control}. The non-diagonal coupling inertial terms were considered unknown, which provides a platform to verify the effect of unmodelled inertial state-dependent forces on control performance. Note that the proposed controller does not require such parametric knowledge.

{\color{black}The control parameters used for the proposed controller during the experiment are listed in Table \ref{tab:control_parameters}.}
\begin{table}[htbp]
\centering
\caption{\color{black} Design Parameters for the Proposed Controller}
\begin{tabular}{|p{0.45\textwidth}|}
\hline
\textbf{\color{black} Quadrotor Position Control} \\
\hline
\vspace{0.1mm}
$\bar{\boldsymbol{M}}_{pp} = \boldsymbol I$, $\boldsymbol Q_p = \text{diag}\{1, 1, 1\}$, $\lambda_{p2} = \text{diag}\{1, 1, 2\}$, $\lambda_{p1} = 2 \lambda_{p2}$, $\boldsymbol  \Lambda_p = \text{diag}\{1.5, 1.5, 2.0\}$; $\hat{K}_{p0}(0) = \hat{K}_{p1}(0) = \hat{K}_{p2}(0) = \hat{K}_{p3}(0) = 0.01$, $\zeta_p(0) = 0.1$, $\nu_{p0} = \nu_{p1} = \nu_{p2} = \nu_{p3} = 10.0$, ${\epsilon}_p = 0.0001$, $\varpi_p = 0.1$ \\
\hline
\hline
\textbf{\color{black} Quadrotor Attitude Control} \\
\hline
\vspace{0.1mm}
$\bar{\boldsymbol M}_{qq} = 0.015 \boldsymbol I$, $\boldsymbol Q_q = \text{diag}\{1, 1, 1\}$, $\lambda_{q2} = \text{diag}\{2, 2, 2\}$, $\lambda_{q1} = 2 \lambda_{q2}$, $\boldsymbol \Lambda_q = \text{diag}\{3.5, 3.5, 2.5\}$; $\hat{K}_{q0}(0) = \hat{K}_{q1}(0) = \hat{K}_{q2}(0) = \hat{K}_{q3}(0) = 0.001$, $\zeta_q(0) = 0.01$, $\nu_{q0} = \nu_{q1} = \nu_{q2} = \nu_{q3} = 20.0$, ${\epsilon}_q = 0.0001$, $\varpi_q = 1.0$ \\
\hline
\hline
\textbf{\color{black} Manipulator Control} \\
\hline
\vspace{0.1mm}
$\bar{\boldsymbol M}_{\alpha \alpha} = 0.1 \boldsymbol I$, $\boldsymbol Q_\alpha = \text{diag}\{1, 1\}$, $\lambda_{\alpha 2} = \text{diag}\{1.5, 1.5\}$, $\lambda_{\alpha 1} = 2 \lambda_{\alpha 2}$, $\boldsymbol \Lambda_\alpha = \text{diag}\{1.0, 1.0\}$; $\hat{K}_{\alpha 0}(0) = \hat{K}_{\alpha 1}(0) = \hat{K}_{\alpha 2}(0) = \hat{K}_{\alpha 3}(0) = 0.0001$, $\zeta_\alpha(0) = 0.01$, $\nu_{\alpha 0} = \nu_{\alpha 1} = \nu_{\alpha 2} = \nu_{\alpha 3} = 1.0$, ${\epsilon}_\alpha = 0.0001$, $\varpi_\alpha = 0.1$ \\
\hline
\end{tabular}
\label{tab:control_parameters}
\end{table}

Note that these choice of gain values lead to different evolution of adaptive gains for each sub-dynamics (cf. discussion below Theorem 1) for the proposed adaptive law compared to common adaptive gains for ASMC. The various control variables of ASMC-1 and ASMC-2 are selected as per \cite{kim2016vision} and \cite{liang2022adaptive}, respectively, after accounting for system parameters. 

\subsection{Results and Analysis}
\begin{figure}[!h]
    \includegraphics[width=0.48\textwidth, height=2.5in]{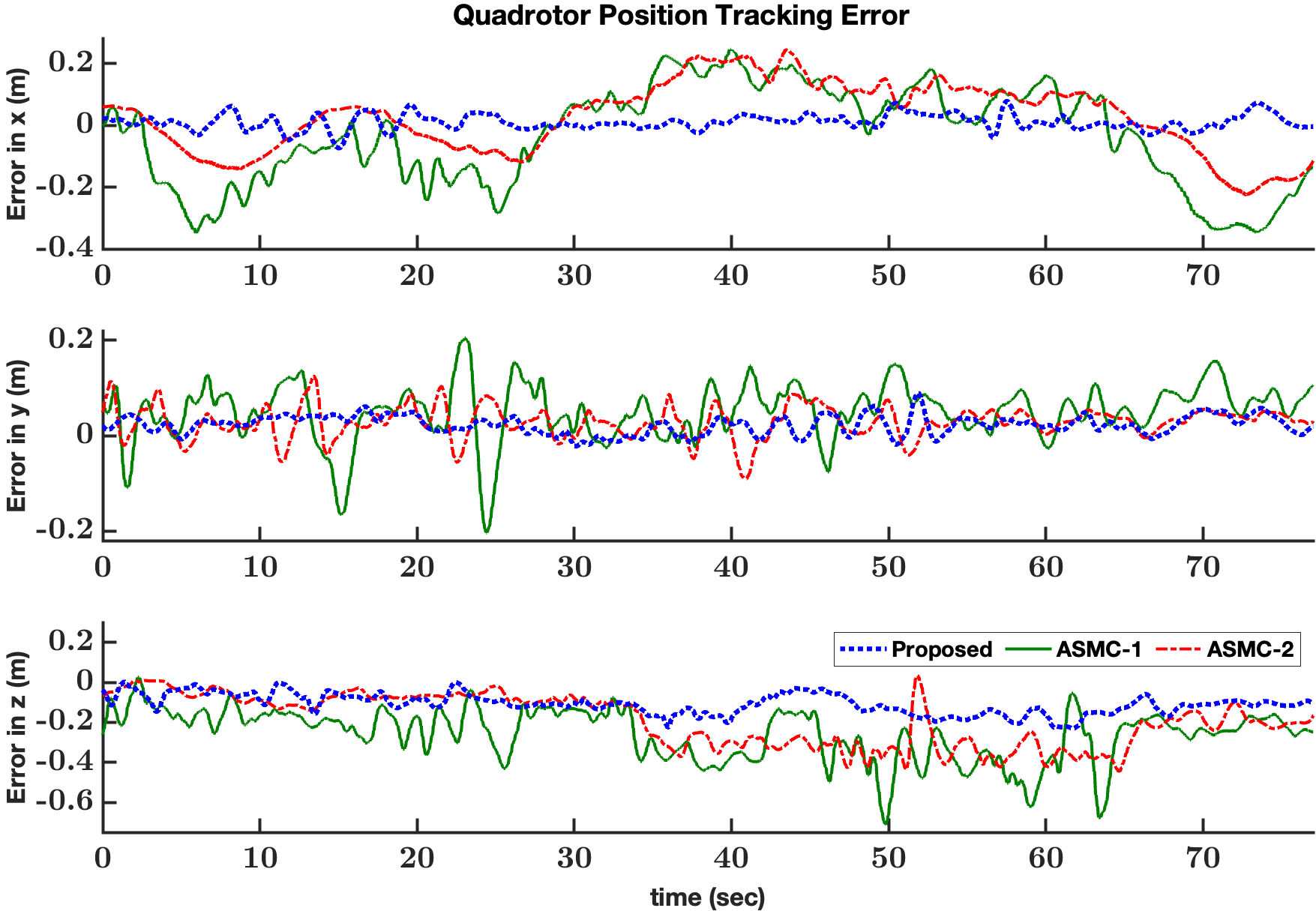}
    \centering
    \caption{\color{black}Comparison of quadrotor position tracking error with various controllers.}
    \label{plot1}
\end{figure}

\begin{figure}[!h]
    \includegraphics[width=0.48\textwidth, height=2.5in]{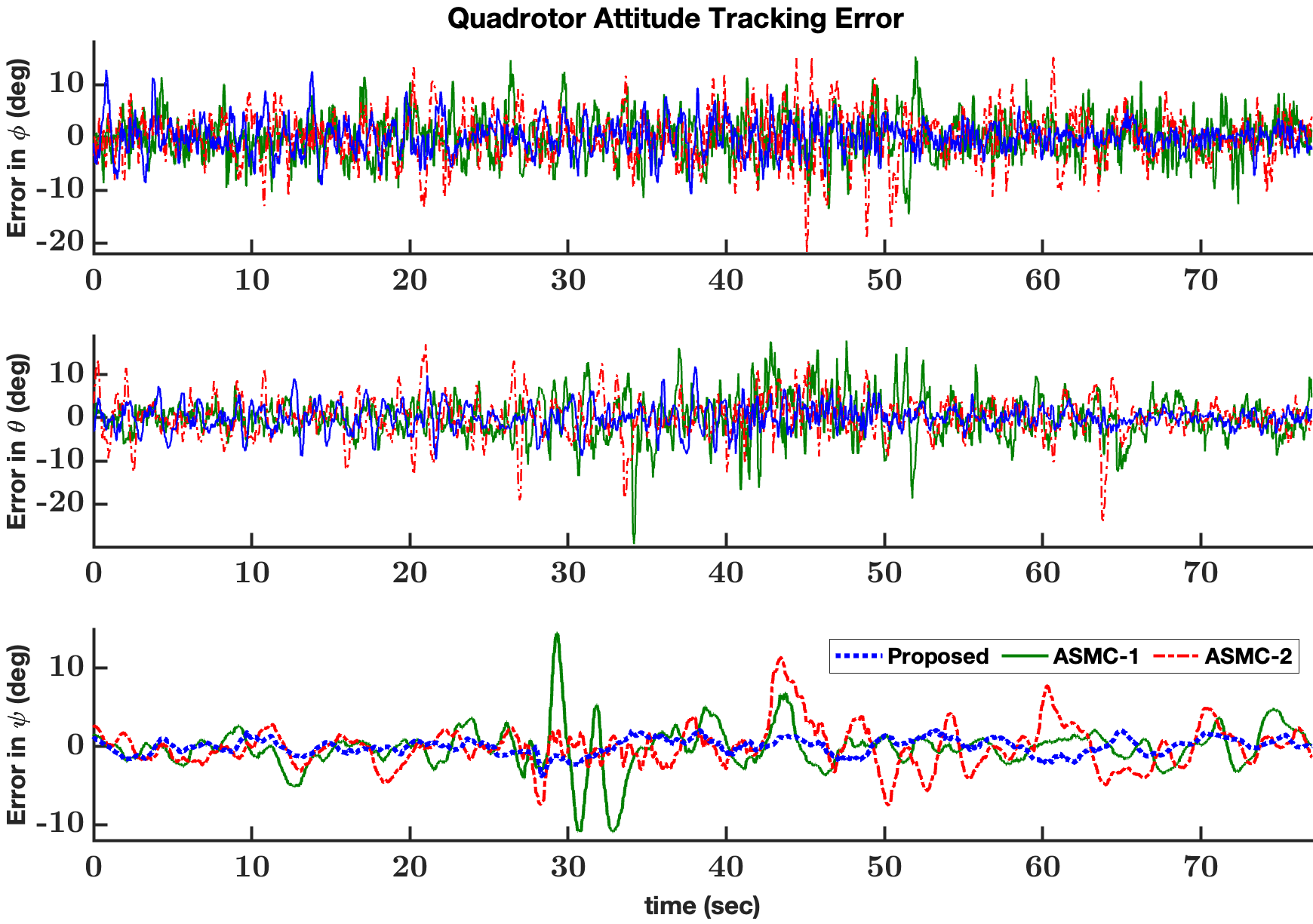}
    \centering
    \caption{\color{black}Comparison of quadrotor attitude tracking error with various controllers.}
    \label{plot3}
\end{figure}

\begin{figure}[!h]
    \includegraphics[width=0.48\textwidth, height=2.5in]{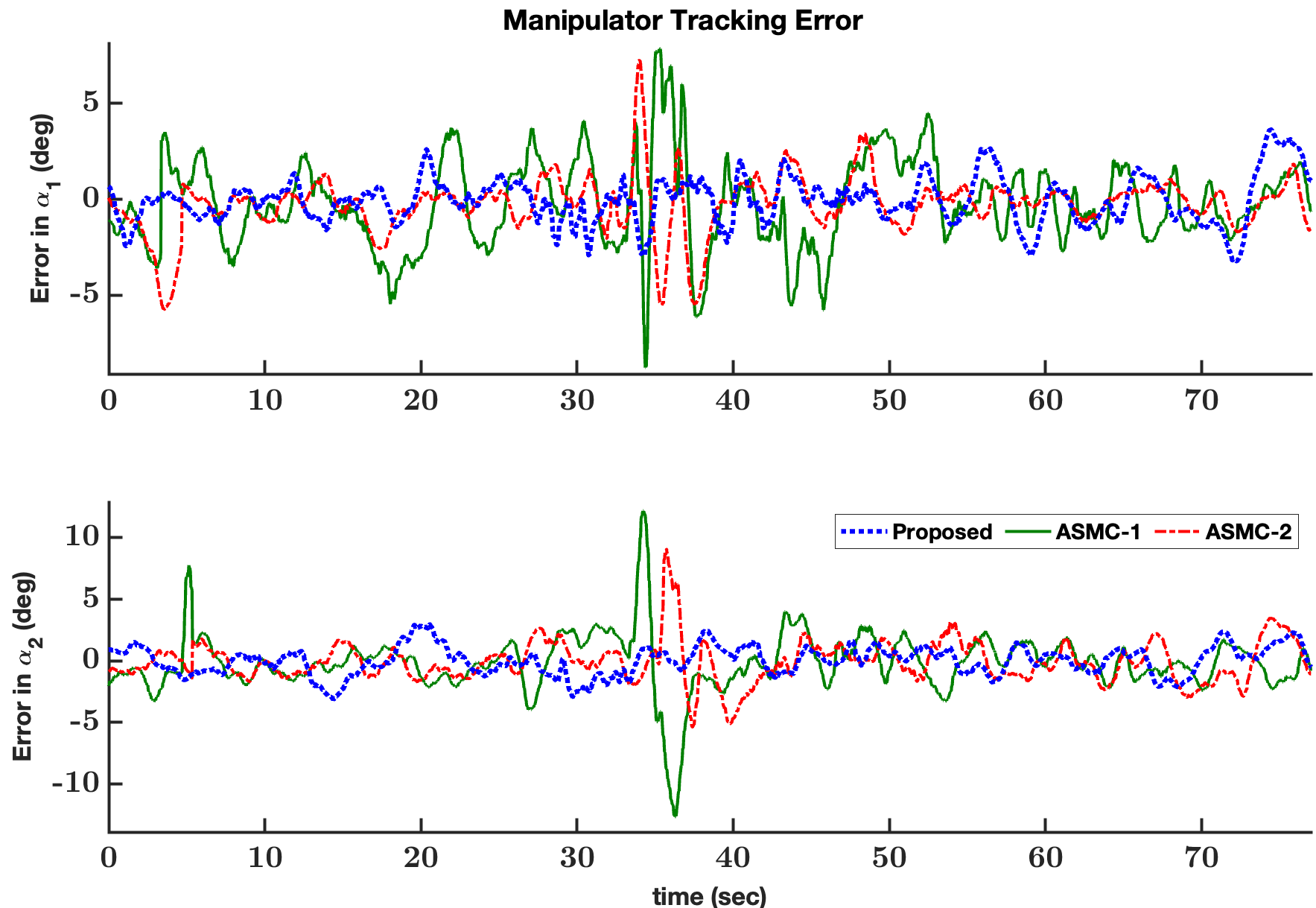}
    \centering
    \caption{\color{black}Comparison of manipulator tracking error with various controllers.}
    \label{plot5}
\end{figure}

\begin{figure}[!h]
    \includegraphics[width=0.48\textwidth, height=2.5in]{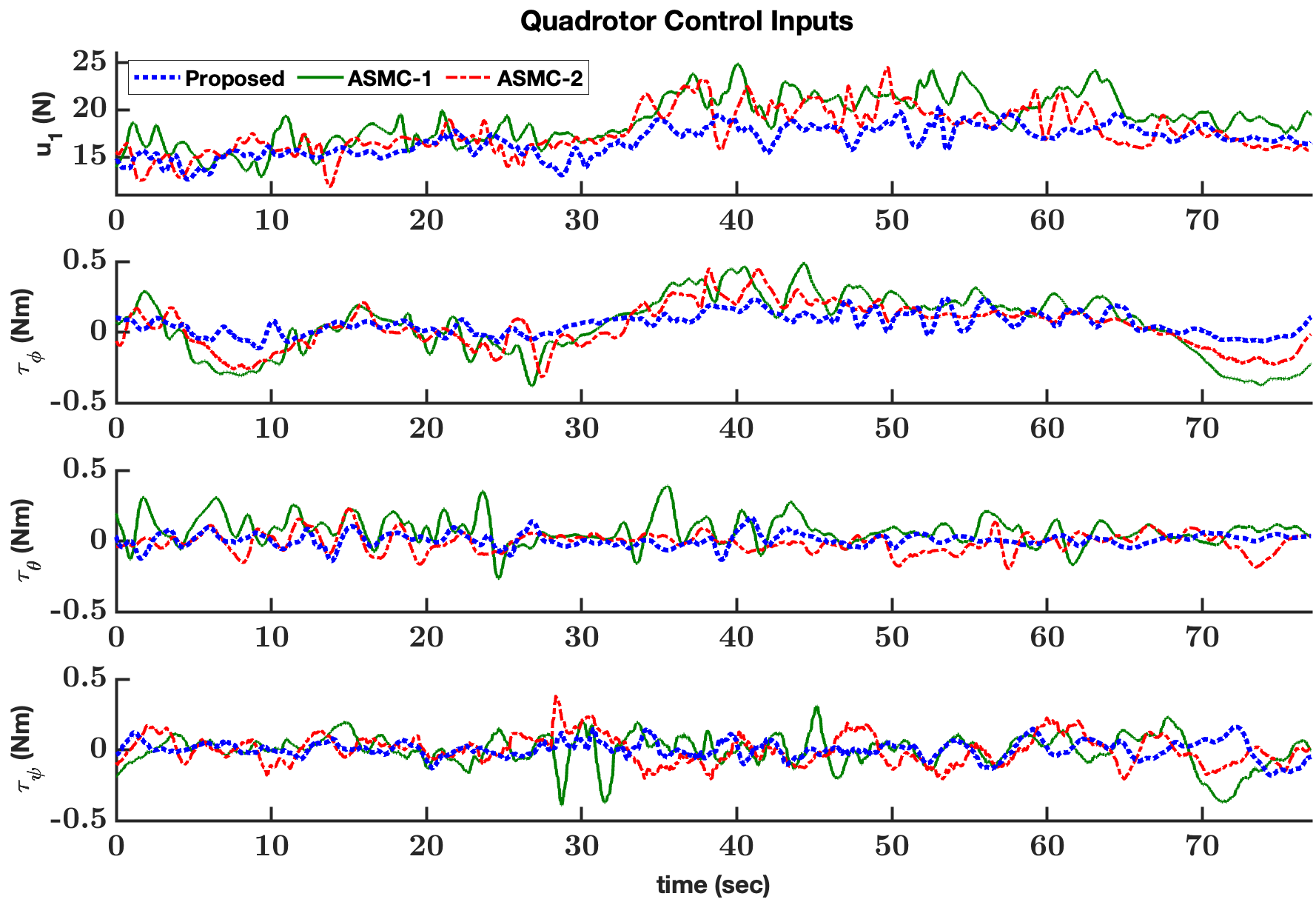}
    \centering
    \caption{\color{black} Comparison of quadrotor control inputs with various controllers.}
    \label{plot6}
\end{figure}

\begin{figure}[!h]
    \includegraphics[width=0.48\textwidth, height=2.5in]{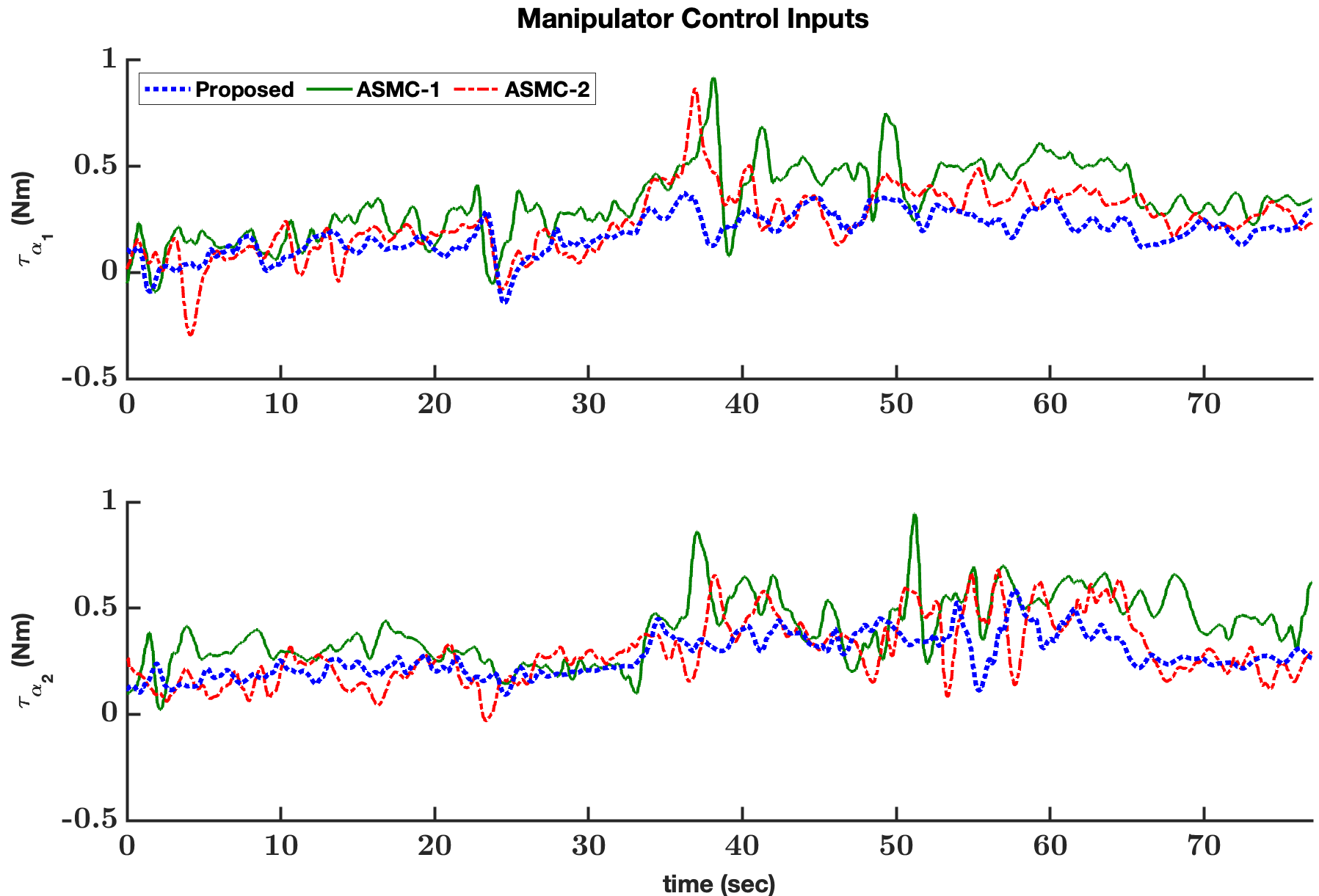}
    \centering
    \caption{\color{black} Comparison of manipulator control inputs with various controllers.}
    \label{plot7}
\end{figure}

The performance of the various controllers are shown in {\color{black}Figs. \ref{plot1}-\ref{plot7}}.  It is evident from these plots that the performance of ASMC-1 and ASMC-2 degraded in $x$ direction (cf. Fig. \ref{plot1}) as manipulator begins extending towards the payload for object retrieval (approx. {\color{black}$2$ sec $<t< 30$ sec}) and similarly, for the duration when manipulator is orienting in opposite direction (approx. {\color{black}$35$ sec $<t< 65$ sec}) to drop the payload. {\color{black}This occurs because both ASMC-1 and ASMC-2 are not designed to tackle unknown inertial coupling forces among quadrotor position, attitude, and manipulator sub-dynamics (cf. Remark 2).} In contrast, the proposed adaptive law is specifically crafted to manage state-dependent uncertainties, including the inertial coupling forces. 

Further, {\color{black}abrupt spikes in the angular position tracking error plots of ASMC-1 and ASMC-2 (cf. Fig. \ref{plot5}) indicate a significant degradation in control performance immediately when the payload is picked up around {\color{black}$t=35$ sec}; consequently, the altitude error (cf. $z$ direction in Fig. \ref{plot1}) also increases for ASMC-1 and ASMC-2 after picking the payload ({\color{black}$t > 35$ sec}) which suggests that the existing controllers cannot handle sudden state-dependent dynamics variations; whereas, the proposed controller leverages the advantages of modular adaptive laws to address these challenges effectively.}

It is crucial to note that achieving precise position tracking for both the quadrotor and the manipulator's angular position is imperative for the successful pickup of the payload. Eventually, the lower Root Mean Square (RMS) errors exhibited by the proposed controller in Tables \ref{table:error_p}-\ref{table:error_alpha} signify its proficient handling of dynamic uncertainties compared to the state of the art. ASMC-2 performed better than ASMC-1 as it can partially tackle state-dependent uncertainties compared to the later.

\begin{table}[!t]
\renewcommand{\arraystretch}{1.1}
\caption{{Quadrotor Position {\color{black}tracking} performance comparison}}
\label{table:error_p}
		\centering
{
{	\begin{tabular}{c c c c c c c}
		\hline
		\hline
		Controller & \multicolumn{3}{c}{RMS position error (m)} & \multicolumn{3}{c}{Performance Degradation ($\%$)}  \\ \cline{1-7}
		 & $x$ & $y$  & $z$  & $x$ & $y$  & $z$  \\
		 \hline
		ASMC-1 & 0.15 & 0.08  & 0.31 &  66.6 &  75.0 & 70.9  \\
		\hline
		ASMC-2 & 0.09 & 0.05  & 0.22 & 44.4 &  60.0 & 59.0 \\
		\hline
		Proposed & 0.05 & 0.02  & 0.09 & - & -  & -  \\
		 \hline
		\hline
\end{tabular}}}
\end{table}

\begin{table}[!t]
\renewcommand{\arraystretch}{1.1}
\caption{{Quadrotor Attitude tracking performance comparison}}
\label{table:error_q}
		\centering
{
{	\begin{tabular}{c c c c c c c}
		\hline
		\hline
		Controller & \multicolumn{3}{c}{RMS attitude error (deg)} & \multicolumn{3}{c}{Performance Degradation ($\%$)}  \\ \cline{1-7}
		 & $\phi$ & $\theta$  & $\psi$  & ${\phi}$ & ${\theta}$  & ${\psi}$  \\
		 \hline
		ASMC-1 & 5.98 & 5.33  & 8.26 & 47.9  & 47.2  & 63.5  \\
		\hline
		ASMC-2 & 4.83 & 4.74  & 5.04 & 35.6 &  40.7 & 40.2 \\
		\hline
		Proposed & 3.11 & 2.81  & 3.01 & - &  - &  - \\
		 \hline
		\hline
\end{tabular}}}
\end{table}

\begin{table}[!t]
\renewcommand{\arraystretch}{1.1}
\caption{{Arm  tracking performance comparison}}
\label{table:error_alpha}
		\centering
{
{	\begin{tabular}{c c c c c}
		\hline
		\hline
		Controller & \multicolumn{2}{c}{RMS  error (deg)} & \multicolumn{2}{c}{Performance Degradation ($\%$)}  \\ \cline{1-5}
		 & $\alpha_1$ & $\alpha_2$  & ${\alpha_1}$ & ${\alpha_2}$    \\
		 \hline

		ASMC-1 & 2.29 & 2.34  & 48.0 & 50.4  \\
		\hline
		ASMC-2 & 1.81 & 1.73  & 34.2 & 32.9 \\
		\hline
		Proposed & 1.19 & 1.16  & - & - \\
		 \hline
		\hline
\end{tabular}}}
\end{table}

\section{Conclusion}
A novel modular adaptive control framework for UAMs was proposed to handle the challenges of unknown inertial dynamic terms and of unknown state-dependent coupling forces. The proposed control method not only contributes to overcome the state-dependent uncertainties in UAMs but also enables independent tuning of adaptive gains to enhance the flexibility and the performance.
The closed-loop stability was verified analytically and real-time experiments validate its superiority over state-of-the-art methods, denoting a significant enhancement for UAM control. 

\section*{Appendix: Proof of Theorem 1}
Modeling the adaptive laws (\ref{adaptive_law_p}), (\ref{adaptive_law_q}) and (\ref{adaptive_law_alpha}) as linear time-varying systems, and using their analytical solutions from positive initial conditions, it can be verified
that $\hat{K}_{ji}\geq 0$ $\forall t>0$ $i=0,\cdots,3, j= p,q, \alpha$ and $\exists \bar{\zeta}_j,  \underline{\zeta}_j \in \mathbb{R}^{+}$ such that
\begin{subequations}\label{bound}
\begin{align} 
0 < \underline{\zeta}_p \leq & \zeta_p (t) \leq \bar{\zeta}_p, ~~ \forall t > 0, \label{bound_p} \\
0 < \underline{\zeta}_q \leq & \zeta_q (t) \leq \bar{\zeta}_q, ~~ \forall t > 0, \label{bound_q}\\
0 < \underline{\zeta}_\alpha \leq & \zeta_\alpha (t) \leq \bar{\zeta}_\alpha, ~~ \forall t > 0 \label{bound_alpha}.
\end{align}
\end{subequations}
Stability is analyzed using the following Lyapunov function:
\begin{align}
V = V_{p} + V_{q} + V_{\alpha}
\label{V}
\end{align}
\begin{align}
\text{with}~V_{p} & = \frac{1}{2} \xi_p^{\top} \boldsymbol P_p \xi_p +  \sum_{i=0}^{3}\frac{(\hat{K}_{pi} -K_{pi}^{*})^2}{2}+\frac{\zeta_p}{\underline{\zeta}_p}, \label{lyap_p}\\
V_{q} & = \frac{1}{2} \xi_q^{\top} \boldsymbol P_q \xi_q +  \sum_{i=0}^{3}\frac{(\hat{K}_{qi} -K_{qi}^{*})^2}{2}+\frac{\zeta_q}{\underline{\zeta}_q}. \label{lyap_q}\\
V_{\alpha} & = \frac{1}{2} \xi_\alpha^{\top} \boldsymbol P_\alpha \xi_\alpha +  \sum_{i=0}^{3}\frac{(\hat{K}_{{\alpha}i} -K_{\alpha i}^{*})^2}{2}+\frac{\zeta_\alpha}{\underline{\zeta}_\alpha}. \label{lyap_alpha}
\end{align}
{\color{black} Taking $\dot{\xi}_j = [e_j^{{\top}}~\dot{e}_j^{\top}]^{\top}$ the standard state-space representations} of the error dynamics in (\ref{err1}), (\ref{err2}) and (\ref{err3}) yield
\begin{subequations} \label{error_dynamics}
\begin{align}
\dot{\xi}_p = & \boldsymbol A_p \xi_p + \boldsymbol B(\sigma_p - \Delta \tau_p), \label{error_dynamics_p}\\
\dot{\xi}_q = & \boldsymbol A_q \xi_q + \boldsymbol B(\sigma_q - \Delta \tau_q), \label{error_dynamics_q}\\
\dot{\xi}_\alpha = & \boldsymbol A_\alpha \xi_\alpha + \boldsymbol B(\sigma_\alpha - \Delta \tau_\alpha) \label{error_dynamics_alpha}.
\end{align}
\end{subequations}
For ease of analysis, we first determine $\dot{V}_p$, $\dot{V}_q$ and $\dot{V}_\alpha$ and then combine them to determine the overall closed-loop stability. In the following, for compactness of notation, we may use the notation $j = p,q,\alpha$ to represent the three terms in the error dynamics or adaptive laws. The process is as follows: 
\subsection{\textbf{Analysis of $\dot{V}_p$}}
 \textbf{Scenario (i): $||r_p|| \geq  \varpi_p$} \\
 Using (\ref{ct1}), (\ref{up_bound_p})-(\ref{adaptive_law_p}),  (\ref{error_dynamics_p}), the fact $\zeta_p>0$ from (\ref{bound_p}) and the Lyapunov equation $\boldsymbol A^{\top}_p \boldsymbol P_p + \boldsymbol P_p A_p = -\boldsymbol Q_p$, the time derivative of $V_p$ yields
 \begin{align} \label{subs_1p}
\dot{V}_p &= \frac{1}{2}\xi_p^{\top}(\boldsymbol A_p^{\top}\boldsymbol P_p + \boldsymbol P_p \boldsymbol A_p)\xi_p + r_p^{\top}\left( \sigma_p - \rho_p \frac{r_p}{||r_p||} \right) \nonumber \\ & + \sum_{i=0}^{3}(\hat{K}_{pi} - K_{pi}^{*}) \dot{\hat{K}}_{pi} + \frac{\dot{\zeta_p}}{\underline{\zeta}_p} \nonumber \\
& \leq - \frac{1}{2}\xi_p^{\top} \boldsymbol Q_p \xi_p + ||\sigma_p|| ||r_p|| - \rho_p ||r_p||  \nonumber  +  \sum_{i=0}^{3}(\hat{K}_{pi} - K_{pi}^{*}) \dot{\hat{K}}_{pi}    \nonumber \\ 
& \leq - \frac{1}{2}\xi_p^{\top} \boldsymbol Q_p \xi_p - \sum_{i=0}^{2}(\hat{K}_{pi} - K_{pi}^{*})||\xi||^i ||r_p||  \nonumber \\ 
& - ( \hat{K}_{p3} - K_{p3}^{*} )||\ddot{\chi}|| ||r_p|| +  \sum_{i=0}^{3}(\hat{K}_{pi} - K_{pi}^{*}) \dot{\hat{K}}_{pi}.
 \end{align}
 
From (\ref{adaptive_law_p}) we have
 \begin{align} \label{subs_2p}
 &\sum_{i=0}^{3}(\hat{K}_{pi} - K_{pi}^{*}) \dot{\hat{K}}_{pi} = \sum_{i=0}^{2}(\hat{K}_{pi} - K_{pi}^{*}) (||r_p|| ||\xi||^i - \nu_{pi} \hat{K}_{pi})
 \nonumber \\ 
 & + 
 (\hat{K}_{p3} - K_{p3}^{*})(||r_p|| ||\ddot{\chi}|| - \nu_{p3} \hat{K}_{p3})
 \nonumber \\ 
 &= \sum_{i=0}^{2}(\hat{K}_{pi} - K_{pi}^{*})||\xi||^i ||r_p|| + (\hat{K}_{p3} - K_{p3}^{*})||\ddot{\chi}|| ||r_p||  
 \nonumber \\ & + 
 \sum_{i=0}^{3}(\nu_{pi} \hat{K}_{pi} K_{pi}^{*} - \nu_{pi} \hat{K}_{pi}^2).  
 \end{align}
 One can verify that
 \begin{equation}
 (\nu_{pi} \hat{K}_{pi} K_{pi}^{*} - \nu_{pi} \hat{K}_{pi}^2) \leq - \frac{\nu_{pi}}{2}  \left((\hat{K}_{pi} -  K_{pi}^*)^2 - {K_{pi}^*}^2\right). \label{ineq}
 \end{equation}
 Substituting (\ref{subs_2p})-{ \color{black}(\ref{ineq})} into (\ref{subs_1p}) yields
 \begin{align} \label{subs_4p}
\dot{V}_p& \leq -\frac{1}{2}\lambda_{\min}(\boldsymbol Q_p)||\xi_p||^2 \nonumber \\
&-\sum \limits_{i=0}^{2} \frac{\nu_{pi}}{2} \left((\hat{K}_{pi} -  K_{pi}^*)^2 - {K_{pi}^*}^2\right).
 \end{align}
The definition of $V_p$ as in (\ref{lyap_p}) yields
 \begin{align} \label{subs_5p}
V_p \leq \frac{1}{2}\lambda_{\max}(\boldsymbol P_p)||\xi_p||^2 + \sum_{i=0}^{3}\frac{(\hat{K}_{pi} - {K_{pi}^{*}})^2}{2}+\frac{\bar{\zeta}_p}{\underline{\zeta}_p}    
 \end{align}
Using (\ref{subs_5p}), the condition (\ref{subs_4p}) is further simplified to
 \begin{align} \label{subs_6p}
\dot{V}_p \leq -\varrho_p V_p + \varrho_p \frac{\bar{\zeta}_p}{\underline{\zeta}_p}  +  \frac{1}{2}\sum \limits_{i=0}^{3} \nu_{pi} {K^*_{pi}}^2
 \end{align}
 where $\varrho_p \triangleq  \frac{\min( \lambda_{\min}(\boldsymbol Q_p),{\nu_{pi}} )}{\max(\lambda_{\max}(\boldsymbol P_p) ,1)} >0$.
 
 \textbf{Scenario (ii): $||r_p|| <  \varpi_p$} \\
 In this scenario we have
  \begin{align} 
\dot{V}_p 
& \leq - \frac{1}{2}\xi_p^{\top} \boldsymbol Q_p \xi_p + ||\sigma_p|| ||r_p|| - \rho_p\frac{ ||r_p||^2}{\varpi_p}  \nonumber \\ & +  \sum_{i=0}^{3}(\hat{K}_{pi} - K_{pi}^{*}) \dot{\hat{K}}_{pi}  + \frac{\dot{\zeta_p}}{\underline{\zeta_p}}  \nonumber \\ 
& \leq  -\frac{1}{2}\lambda_{\min}(\boldsymbol Q_p)||\xi_p||^2 + ||\sigma_p|| ||r_p||   \nonumber \\
&+  \sum_{i=0}^{3}(\hat{K}_{pi} - K_{pi}^{*}) \dot{\hat{K}}_{pi}+ \frac{\dot{\zeta_p}}{\underline{\zeta_p}} \nonumber\\
&\leq  -\frac{1}{2}\lambda_{\min}(\boldsymbol Q_p)||\xi||^2 +
\sum_{i=0}^{2}\hat{K}_{pi} ||\xi||^i ||r_p|| + \hat{K}_{p3} ||\ddot{\chi}|| ||r_p|| \nonumber\\
&-\sum \limits_{i=0}^{2} \frac{\nu_{pi}}{2} \left((\hat{K}_{pi} -  K_{pi}^*)^2 - {K_{pi}^*}^2\right)    + \frac{\dot{\zeta_p}}{\underline{\zeta_p}}. \label{new_2}
 \end{align}

The adaptive law (\ref{zeta1}) and relation (\ref{bound_p}) lead to
 \begin{align} 
 \label{subs_3p}
\frac{\dot{\zeta_p}}{\underline{\zeta}_p} &= -\left(1 + \hat{K}_{p3} ||\ddot{\chi}|| ||r_p|| + \sum_{i=0}^{2}\hat{K}_{pi} ||\xi||^i ||r_p||\right)  \frac{{\zeta_p}}{\underline{\zeta}_p} + \frac{{{\epsilon}_p}}{\underline{\zeta}_p}   \nonumber \\
& \leq  -\hat{K}_{p3} ||\ddot{\chi}|| ||r_p|| - \sum_{i=0}^{2}\hat{K}_{pi} ||\xi||^i ||r_p|| + \frac{{{\epsilon}_p}}{\underline{\zeta}_p}.
 \end{align}
Substituting (\ref{subs_3p}) into (\ref{new_2}) and using (\ref{subs_5p}), $\dot{V}_p$ is simplified to 
\begin{align} 
\dot{V}_p &\leq -\varrho_p V_p + \varrho_p \frac{\bar{\zeta}_p}{\underline{\zeta}_p} + \frac{{{\epsilon}_p}}{\underline{\zeta}_p}  +  \frac{1}{2}\sum \limits_{i=0}^{3} \nu_{pi} {K^*_{pi}}^2. \label{new2}
\end{align}

\subsection{\textbf{Analysis of $\dot{V}_q$}}
 \textbf{Scenario (i): $||r_q|| \geq  \varpi_q$} \\
Using (\ref{ct2}), (\ref{up_bound_q})-(\ref{adaptive_law_q}),  (\ref{error_dynamics_q}), the Lyapunov equation $\boldsymbol A^{\top}_q \boldsymbol P_q + \boldsymbol P_q \boldsymbol A_q = -\boldsymbol Q_q$ and following the derivations of $\dot{V}_p$, we have
\begin{align} \label{subs_1q}
\dot{V}_q & \leq -\frac{1}{2}\lambda_{\min}(\boldsymbol Q_q)||\xi_q||^2 + \frac{{{\epsilon}_q}}{\underline{\zeta}_q} \nonumber \\&-
\sum \limits_{i=0}^{2} \frac{\nu_{qi}}{2} \left((\hat{K}_{qi} -  K_{qi}^*)^2 - {K_{qi}^*}^2\right).
 \end{align}
The definition of $V_q$ as in (\ref{lyap_q}) yields
 \begin{align} \label{subs_2q}
V_q \leq \frac{1}{2}\lambda_{\max}(\boldsymbol P_q)||\xi_q||^2 + \sum_{i=0}^{3}\frac{(\hat{K}_{qi} - {K_{qi}^{*}})^2}{2}+\frac{\bar{\zeta}_q}{\underline{\zeta}_q}.    
 \end{align}
Using (\ref{subs_2q}), (\ref{subs_1q}) is simplified to
 \begin{align} \label{subs_3q}
\dot{V}_q \leq -\varrho_q V_q + \varrho_q \frac{\bar{\zeta}_q}{\underline{\zeta}_q}  +  \frac{1}{2}\sum \limits_{i=0}^{3} \nu_{qi} {K^*_{qi}}^2
 \end{align}
 where $\varrho_q \triangleq  \frac{\min( \lambda_{\min}(\boldsymbol Q_q),{\nu_{qi}} )}{\max(\lambda_{\max}(\boldsymbol P_q) ,1)} >0$. 
 
  \textbf{Scenario (ii): $||r_q|| <  \varpi_q$} \\
Following the steps for $\dot{V}_p$ for this scenario, we have
  \begin{align} 
\dot{V}_q 
& \leq -\frac{1}{2}\lambda_{\min}(\boldsymbol Q_q)||\xi_q||^2 +
\sum_{i=0}^{2}\hat{K}_{qi} ||\xi||^i ||r_q|| + \hat{K}_{q3} ||\ddot{\chi}|| ||r_q||  \nonumber\\
& -\sum \limits_{i=0}^{2} \frac{\nu_{qi}}{2} \left((\hat{K}_{qi} -  K_{qi}^*)^2 - {K_{qi}^*}^2\right) + \frac{\dot{\zeta_q}}{\underline{\zeta_q}}\nonumber\\
&\leq -\varrho_q V_q + \varrho_q \frac{\bar{\zeta}_q}{\underline{\zeta}_q} + \frac{{{\epsilon}_q}}{\underline{\zeta}_q}  +  \frac{1}{2}\sum \limits_{i=0}^{3} \nu_{qi} {K^*_{qi}}^2. \label{new3}
\end{align}
 
 \subsection{\textbf{Analysis of $\dot{V}_\alpha$}}
 \textbf{Scenario (i): $||r_\alpha|| \geq  \varpi_\alpha$} \\
Using (\ref{ct3}), (\ref{up_bound_alpha})-(\ref{adaptive_law_alpha}),  (\ref{error_dynamics_alpha}), the Lyapunov equation $\boldsymbol A^{\top}_\alpha \boldsymbol P_\alpha + \boldsymbol P_\alpha \boldsymbol A_\alpha = - \boldsymbol Q_\alpha$, and following the similar procedures to derive $\dot{V}_p$ and $\dot{V}_q$, one can derive
  \begin{align} \label{subs_3alpha}
\dot{V}_\alpha \leq -\varrho_\alpha V_\alpha + \varrho_\alpha \frac{\bar{\zeta}_\alpha}{\underline{\zeta}_\alpha}  +  \frac{1}{2}\sum \limits_{i=0}^{3} \nu_{\alpha i} {K^*_{\alpha i}}^2
 \end{align}
 where $\varrho_\alpha \triangleq  \frac{\min( \lambda_{\min}(\boldsymbol Q_\alpha),{\nu_{\alpha i}} )}{\max(\lambda_{\max}(\boldsymbol P_\alpha) ,1)} >0$. 
 
  \textbf{Scenario (ii): $||r_\alpha|| <  \varpi_\alpha$} \\
  In this case, following similar lines of proof for $\dot{V}_p$ and for $\dot{V}_q$, we have
\begin{align} 
\dot{V}_\alpha &\leq -\varrho_\alpha V_\alpha + \varrho_\alpha \frac{\bar{\zeta}_\alpha}{\underline{\zeta}_\alpha} + \frac{{{\epsilon}_\alpha}}{\underline{\zeta}_\alpha}  +  \frac{1}{2}\sum \limits_{i=0}^{3} \nu_{\alpha i} {K^*_{\alpha i}}^2. \label{new4}
\end{align}
\subsection{\textbf{Overall Stability Analysis}}
For the overall stability of $\dot{V}$, we list the various possible cases: \\
\textbf{Case (i): $||r_p|| \geq  \varpi_p, ||r_q|| \geq  \varpi_q, ||r_\alpha|| \geq  \varpi_\alpha$} \\
\textbf{Case (ii): $||r_p|| <  \varpi_p, ||r_q|| <  \varpi_q, ||r_\alpha|| <  \varpi_\alpha$} \\
\textbf{Case (iii): $||r_p|| <  \varpi_p, ||r_q|| \geq  \varpi_q, ||r_\alpha|| \geq  \varpi_\alpha$} \\
\textbf{Case (iv): $||r_p|| \geq  \varpi_p, ||r_q|| <  \varpi_q, ||r_\alpha|| \geq \varpi_\alpha$} \\
\textbf{Case (v): $||r_p|| \geq  \varpi_p, ||r_q|| \geq  \varpi_q, ||r_\alpha|| <  \varpi_\alpha$} \\
\textbf{Case (vi): $||r_p|| <  \varpi_p, ||r_q|| <  \varpi_q, ||r_\alpha|| \geq  \varpi_\alpha$} \\
\textbf{Case (vii): $||r_p|| \geq  \varpi_p, ||r_q|| <  \varpi_q, ||r_\alpha|| <  \varpi_\alpha$} \\
\textbf{Case (viii): $||r_p|| <  \varpi_p, ||r_q|| \geq  \varpi_q, ||r_\alpha|| <  \varpi_\alpha$} \\

Observing the results of $\dot{V}_j$, $j=p,q,\alpha$ under various scenarios as in (\ref{subs_6p}), (\ref{new2}), (\ref{subs_3q}), (\ref{new3}) and (\ref{subs_3alpha}), (\ref{new4}) the common time derivative of Lyapunov function ${V}$ for Cases (i)-(viii) from (\ref{V}) is obtained as 
\begin{align}
\dot{V} & \leq -\varrho V + \gamma,  \label{new} \\
\text{where}~\varrho &= \min \{ \varrho_p, \varrho_q, \varrho_\alpha \}, \nonumber \\
\gamma &=\sum \limits_{j=p, q, \alpha}^{} \left(\varrho_j \frac{\bar{\zeta}_j}{\underline{\zeta}_j} + \frac{{{\epsilon}_j}}{\underline{\zeta}_j} +  \frac{1}{2}\sum \limits_{i=0}^{3} \nu_{ji} {K^*_{ji}}^2\right).
\nonumber 
\end{align}
Defining a scalar $\kappa$ as $0<\kappa<\varrho$, (\ref{new}) can be simplified to
\begin{align}
\dot{V} & \leq  -\kappa V - (\varrho - \kappa)V + \gamma.
\end{align}
Further defining a scalar $\mathcal{  B} = \frac{\gamma}{(\varrho - \kappa)}$
 it can be concluded that $\dot{V} (t) < - \kappa V (t)$ when $V (t) \geq \mathcal{ B}$, so that
\begin{align}
    V & \leq \max \{ V(0), \mathcal{  B} \}, ~\forall t \geq 0,
\end{align}
and the closed-loop system remains UUB (cf. UUB definition $4.6$ as in \cite{khalil2002nonlinear}).

\begin{remark}[Role of $\zeta_j$]
\label{rem_zeta}
The boundedness of $||r_j|| \leq \varpi_j, j = p, q, \alpha$ in Scenario (ii) of $\dot{V}_j$ implies boundedness of $||\xi_j||$ but not of $||\ddot{\chi}||$ and of $||\xi||$. Therefore, canceling the terms `($\hat{K}_{j3} ||\ddot{\chi}|| ||r_j|| + \sum_{i=0}^{2}\hat{K}_{ji} ||\xi||^i ||r_j||) $' through the auxiliary gains $\zeta_j$ (cf. $\dot{\zeta}_j$) is necessary to guarantee closed-loop stability.
\end{remark}


\begin{IEEEbiography}[{\includegraphics[width=1in,height=1.25in,clip,keepaspectratio]{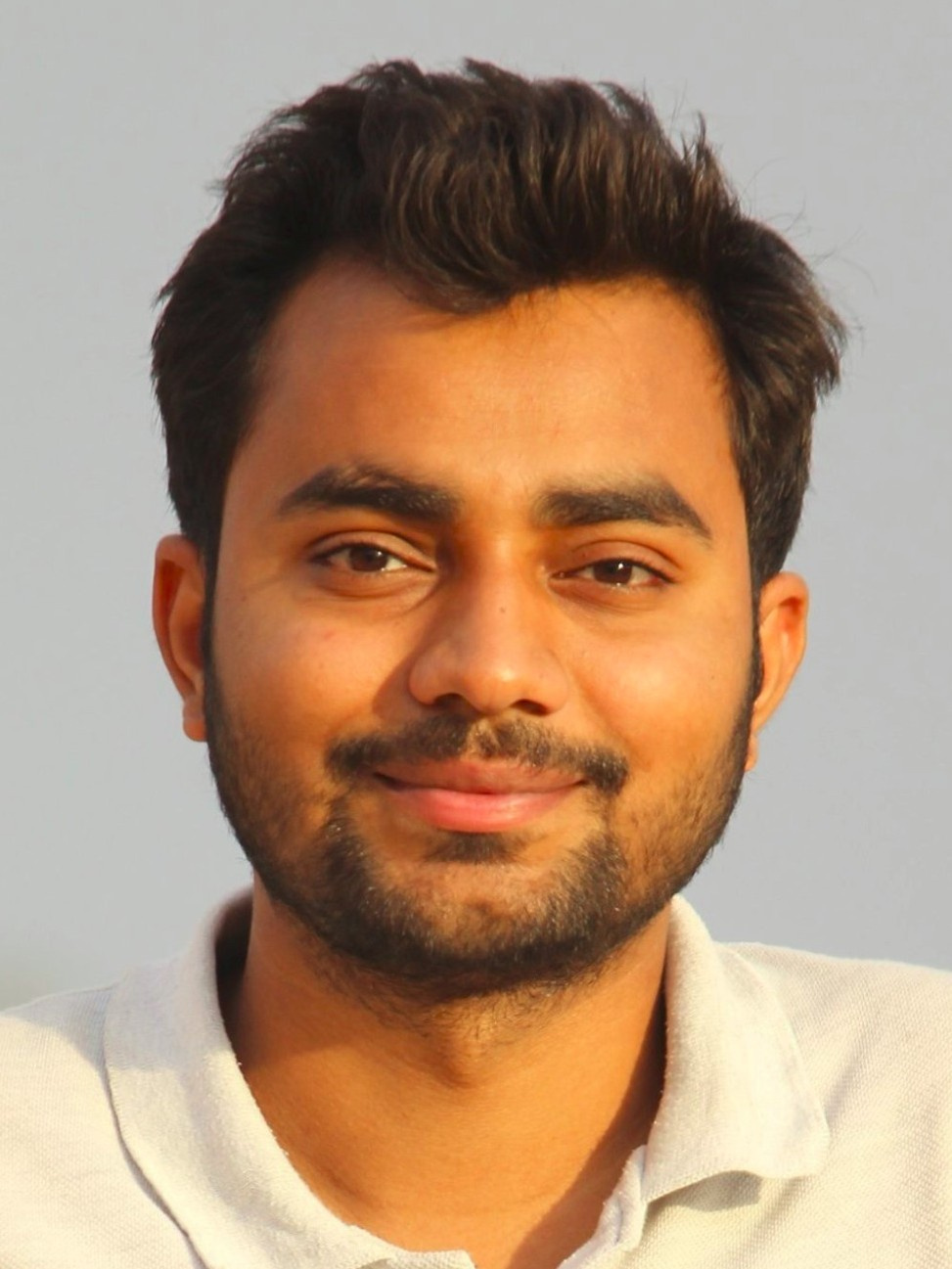}}]{Rishabh Dev Yadav} received his B.Tech degree in Mechanical Engineering from Indian Institute of Information Technology Jabalpur, India in 2019; Master of Science in Computer Science and Engineering by Research from International Institute of Information Technology Hyderabad, India in 2023. He is a Ph.D. student at the Department of Computer Science, The University of Manchester, UK. His research interests include applied adaptive-robust control for Unmanned Aerial Vehicles.  
\end{IEEEbiography}

\begin{IEEEbiography}[{\includegraphics[width=1in,height=1.25in,clip,keepaspectratio]{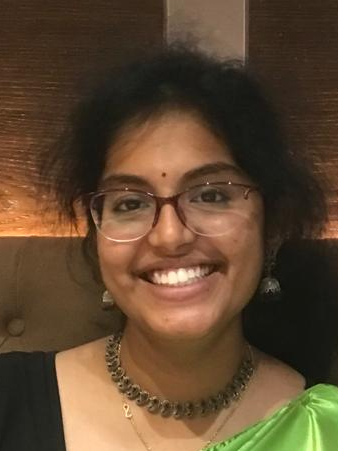}}]{Swati Dantu} received the B.Tech degree in Electronics and Communication Engineering from Jawaharlal Nehru Technological University Hyderabad, India in 2020, MS by Research in Electronics and Communication Engineering from International Institute of Information Technology Hyderabad, India in 2023. She is currently a Ph.D. student at the Czech Technical University in Prague, Czechia.  
Her current research interests include learning based adaptive and robust controls for Unmanned Aerial Vehicles.

\end{IEEEbiography}

\begin{IEEEbiography}[{\includegraphics[width=1in,height=1.25in,clip,keepaspectratio]{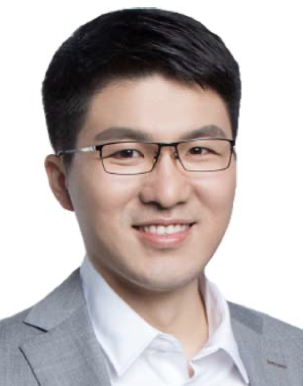}}]{Wei Pan} received the Ph.D. degree in bioengineering from Imperial College London, London, U.K., in 2017. He is currently an Associate Professor in Machine Learning and director of Robotics and Embodied AI Lab (REAL) at the Department of Computer Science and a member of Centre for AI Fundamentals and Centre for Robotics and AI, The University of Manchester, UK. Before that, he was an Assistant Professor in Robot Dynamics at the Department of Cognitive Robotics and co-director of Delft SELF AI Lab, TU Delft, Netherlands and a Project Leader at DJI, China. Dr. Pan is the recipient of Dorothy Hodgkin’s Postgraduate Awards, Microsoft Research Ph.D. Scholarship, and Chinese Government Award for Outstanding Self-financed Students Abroad, Shenzhen Peacock Plan Award. He is the Area Chair or (Senior) Associate Editor of \textit{IEEE Robotics and Automation Letters} (outstanding AE award), \textit{ACM Transactions on Probabilistic Machine Learning}, \textit{Conference on Robot Learning (CoRL)}, \textit{Conference on Learning for Dynamics and Control (L4DC)}, \textit{IEEE International Conference on Robotics and Automation (ICRA)}, \textit{IEEE/RSJ International Conference on Intelligent Robots and Systems (IROS)}.
His research interests include machine learning and control theory with applications in robotics.
\end{IEEEbiography}

\begin{IEEEbiography}[{\includegraphics[width=1in,height=1.25in,clip,keepaspectratio]{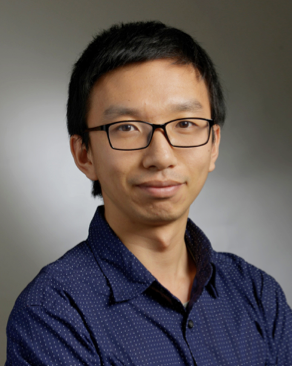}}]{Sihao Sun} received the B.Sc. and M.Sc. degrees in aerospace engineering from Beihang University, Beijing, China, in 2014 and 2017, respectively. In 2020, he received a Ph.D. degree in aerospace engineering from Delft University of Technology, Delft, Netherlands. From 2020 to 2021, he was first a visiting scholar and then a postdoctoral researcher in the Robotics and Perception Group, University of Zurich, Switzerland. Since 2022, he has been a postdoctoral researcher at the Robotics and Mechatronics (RaM) group at the University of Twente. He is now a researcher at the Department of Cognitive Robotics at Delft University of Technology. He is the recipient of Veni grant as part of the Dutch Research Council (NWO) Talent Program, and winner of the Best Paper Award of IEEE Robotics and Automation Letters (RA-L). He is serving as an associate editor of \textit{IEEE Robotics and Automation Letters (RA-L), IEEE/RSJ International Conference on Intelligent Robots and Systems (IROS)}. His research interests include system identification, aerial robotics, and nonlinear control.
\end{IEEEbiography}

\begin{IEEEbiography}[{\includegraphics[width=1in,height=1.25in,clip,keepaspectratio]{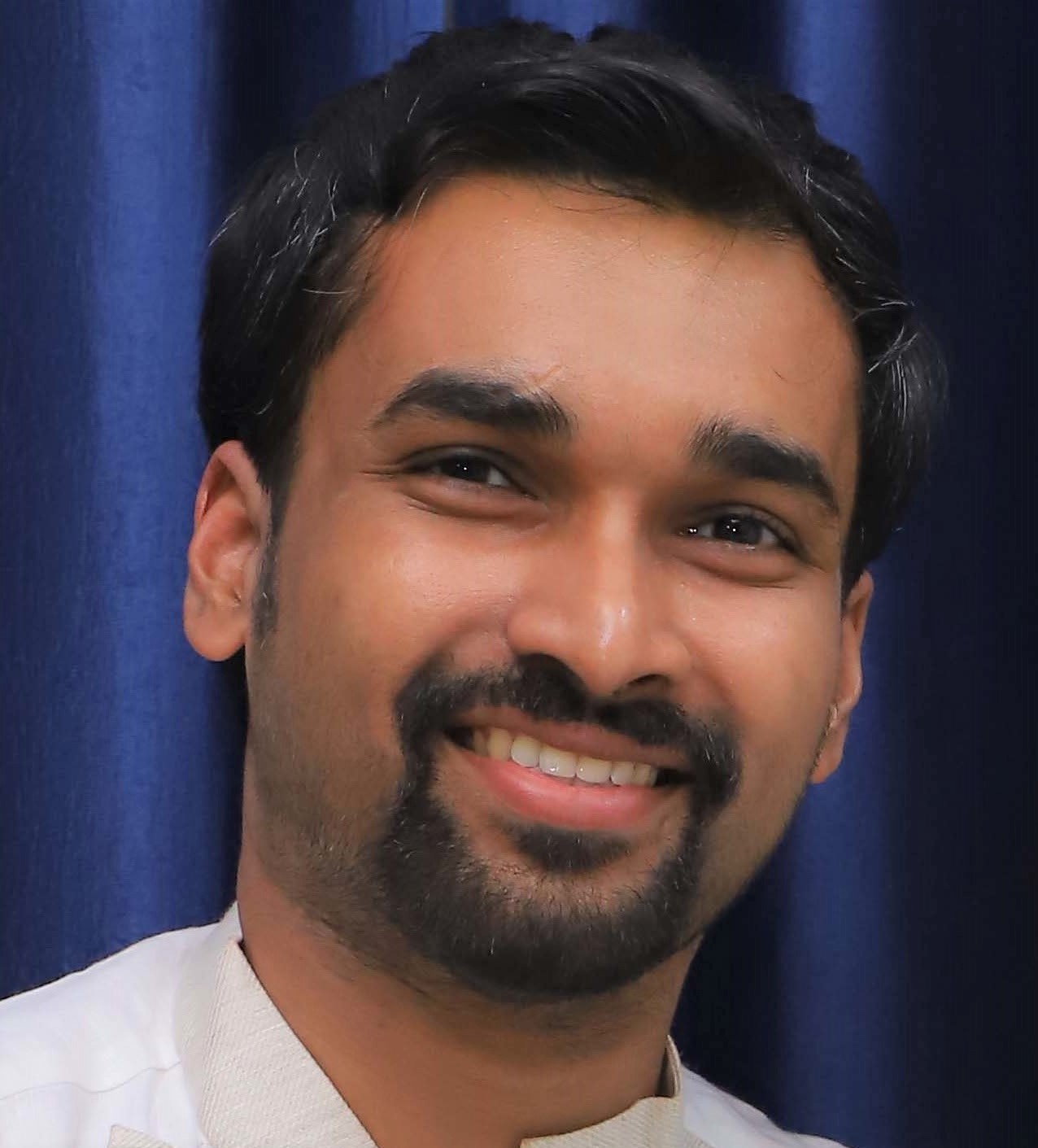}}]{Spandan Roy} (M'18) received his B.Tech degree in Electronics and Communication Engineering from Techno India, West Bengal University of Technology,  India in 2011, M.Tech. degree in Mechatronics from Academy of Scientific and Innovative Research, India in 2013 and Ph.D. degree in Control and Automation from Indian Institute of Technology Delhi, India in 2018. He is currently an assistant professor at the Robotics Research Center, International Institute of Information Technology Hyderabad, India. Previously, he was a postdoctoral researcher in Delft Center for System and Control, Delft University of Technology, The Netherlands. He is a subject editor at \textit{International Journal of Adaptive Control and Signal Processing}. His research interests include adaptive-robust control, switched systems, and its applications in Euler-Lagrange systems.
\end{IEEEbiography}

\begin{IEEEbiography}[{\includegraphics[width=1in,height=1.25in,clip,keepaspectratio]{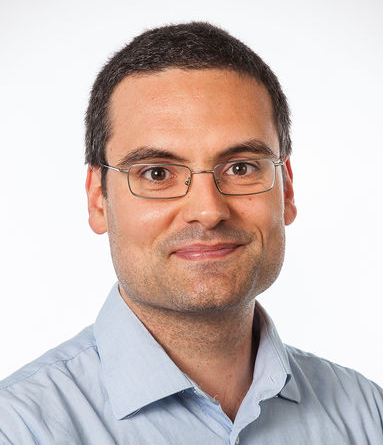}}]
{Simone Baldi} (M'14, SM'19) received the B.Sc. in electrical engineering, and the M.Sc. and Ph.D. in automatic control engineering from University of Florence, Italy, in 2005, 2007, and 2011. He is currently a Professor with the School of Mathematics, Southeast University. Before that, he was an assistant professor at Delft Center for Systems and Control, TU Delft. He was awarded outstanding reviewer for \emph{Applied Energy} (2016) and \emph{Automatica} (2017). He is a subject editor of \emph{Int. Journal of Adaptive Control and Signal Processing}, an associate editor of \emph{IEEE Control Systems Letters} and a technical editor of \emph{IEEE/ASME Trans. on Mechatronics}. His research interests are adaptive and learning systems with applications in unmanned vehicles and smart energy systems.
  
\end{IEEEbiography}

\vfill

\begin{thebibliography}{10}
\providecommand{\url}[1]{#1}
\csname url@samestyle\endcsname
\providecommand{\newblock}{\relax}
\providecommand{\bibinfo}[2]{#2}
\providecommand{\BIBentrySTDinterwordspacing}{\spaceskip=0pt\relax}
\providecommand{\BIBentryALTinterwordstretchfactor}{4}
\providecommand{\BIBentryALTinterwordspacing}{\spaceskip=\fontdimen2\font plus
\BIBentryALTinterwordstretchfactor\fontdimen3\font minus \fontdimen4\font\relax}
\providecommand{\BIBforeignlanguage}[2]{{%
\expandafter\ifx\csname l@#1\endcsname\relax
\typeout{** WARNING: IEEEtran.bst: No hyphenation pattern has been}%
\typeout{** loaded for the language `#1'. Using the pattern for}%
\typeout{** the default language instead.}%
\else
\language=\csname l@#1\endcsname
\fi
#2}}
\providecommand{\BIBdecl}{\relax}
\BIBdecl

\bibitem{fumagalli2014developing}
M.~Fumagalli, R.~Naldi, A.~Macchelli, F.~Forte, A.~Q. Keemink, S.~Stramigioli, R.~Carloni, and L.~Marconi, ``Developing an aerial manipulator prototype: Physical interaction with the environment,'' \emph{IEEE Robotics \& Automation Magazine}, vol.~21, no.~3, pp. 41--50, 2014.

\bibitem{fumagalli2016mechatronic}
M.~Fumagalli, S.~Stramigioli, and R.~Carloni, ``Mechatronic design of a robotic manipulator for unmanned aerial vehicles,'' in \emph{2016 IEEE/RSJ International Conference on Intelligent Robots and Systems (IROS)}.\hskip 1em plus 0.5em minus 0.4em\relax IEEE, 2016, pp. 4843--4848.

\bibitem{mimmo2020robust}
N.~Mimmo, A.~Macchelli, R.~Naldi, and L.~Marconi, ``Robust motion control of aerial manipulators,'' \emph{Annual Reviews in Control}, vol.~49, pp. 230--238, 2020.

\bibitem{loianno2018localization}
G.~Loianno, V.~Spurny, J.~Thomas, T.~Baca, D.~Thakur, D.~Hert, R.~Penicka, T.~Krajnik, A.~Zhou, A.~Cho \emph{et~al.}, ``Localization, grasping, and transportation of magnetic objects by a team of {MAV}s in challenging desert-like environments,'' \emph{IEEE Robotics and Automation Letters}, vol.~3, no.~3, pp. 1576--1583, 2018.

\bibitem{zhong2019practical}
H.~Zhong, Z.~Miao, Y.~Wang, J.~Mao, L.~Li, H.~Zhang, Y.~Chen, and R.~Fierro, ``A practical visual servo control for aerial manipulation using a spherical projection model,'' \emph{IEEE Transactions on Industrial Electronics}, vol.~67, no.~12, pp. 10\,564--10\,574, 2019.

\bibitem{tognon2019truly}
M.~Tognon, H.~A.~T. Ch{\'a}vez, E.~Gasparin, Q.~Sabl{\'e}, D.~Bicego, A.~Mallet, M.~Lany, G.~Santi, B.~Revaz, J.~Cort{\'e}s \emph{et~al.}, ``A truly-redundant aerial manipulator system with application to push-and-slide inspection in industrial plants,'' \emph{IEEE Robotics and Automation Letters}, vol.~4, no.~2, pp. 1846--1851, 2019.

\bibitem{suarez2020benchmarks}
A.~Suarez, V.~M. Vega, M.~Fernandez, G.~Heredia, and A.~Ollero, ``Benchmarks for aerial manipulation,'' \emph{IEEE Robotics and Automation Letters}, vol.~5, no.~2, pp. 2650--2657, 2020.

\bibitem{li2023pseudospectral}
D.~Li, Y.~Li, X.~Liu, B.~Yang, X.~Huang, Y.~Yang, B.~Wang, and S.~Li, ``Pseudospectral convex programming for free-floating space manipulator path planning,'' \emph{Space: Science \& Technology}, vol.~3, p. 0030, 2023.

\bibitem{orsag2017dexterous}
M.~Orsag, C.~Korpela, S.~Bogdan, and P.~Oh, ``Dexterous aerial robots—mobile manipulation using unmanned aerial systems,'' \emph{IEEE Transactions on Robotics}, vol.~33, no.~6, pp. 1453--1466, 2017.

\bibitem{cao2019faster}
L.~Cao, B.~Xiao, M.~Golestani, and D.~Ran, ``Faster fixed-time control of flexible spacecraft attitude stabilization,'' \emph{IEEE Transactions on Industrial Informatics}, vol.~16, no.~2, pp. 1281--1290, 2020.

\bibitem{tilli2021low}
A.~Tilli, E.~Ruggiano, A.~Bosso, and A.~Samor{\`\i}, ``Low-input accurate periodic motion of an underactuated mechanism: Mass distribution and nonlinear spring shaping,'' in \emph{2021 IEEE/ASME International Conference on Advanced Intelligent Mechatronics (AIM)}.\hskip 1em plus 0.5em minus 0.4em\relax IEEE, 2021, pp. 292--299.

\bibitem{ruggiero2018aerial}
F.~Ruggiero, V.~Lippiello, and A.~Ollero, ``Aerial manipulation: A literature review,'' \emph{IEEE Robotics and Automation Letters}, vol.~3, no.~3, pp. 1957--1964, 2018.

\bibitem{xiao2021prescribed}
B.~Xiao, X.~Wu, L.~Cao, and X.~Hu, ``Prescribed time attitude tracking control of spacecraft with arbitrary disturbance,'' \emph{IEEE Transactions on Aerospace and Electronic Systems}, vol.~58, no.~3, pp. 2531--2540, 2022.

\bibitem{ollero2021past}
A.~Ollero, M.~Tognon, A.~Suarez, D.~Lee, and A.~Franchi, ``Past, present, and future of aerial robotic manipulators,'' \emph{IEEE Transactions on Robotics}, 2021.

\bibitem{jimenez2013control}
A.~E. Jimenez-Cano, J.~Martin, G.~Heredia, A.~Ollero, and R.~Cano, ``Control of an aerial robot with multi-link arm for assembly tasks,'' in \emph{2013 IEEE International Conference on Robotics and Automation}.\hskip 1em plus 0.5em minus 0.4em\relax IEEE, 2013, pp. 4916--4921.

\bibitem{ruggiero2015multilayer}
F.~Ruggiero, M.~A. Trujillo, R.~Cano, H.~Ascorbe, A.~Viguria, C.~Per{\'e}z, V.~Lippiello, A.~Ollero, and B.~Siciliano, ``A multilayer control for multirotor {UAV}s equipped with a servo robot arm,'' in \emph{2015 IEEE international conference on robotics and automation (ICRA)}.\hskip 1em plus 0.5em minus 0.4em\relax IEEE, 2015, pp. 4014--4020.

\bibitem{suarez2018physical}
A.~Suarez, G.~Heredia, and A.~Ollero, ``Physical-virtual impedance control in ultralightweight and compliant dual-arm aerial manipulators,'' \emph{IEEE Robotics and Automation Letters}, vol.~3, no.~3, pp. 2553--2560, 2018.

\bibitem{kim2018cooperative}
S.~Kim, H.~Seo, J.~Shin, and H.~J. Kim, ``Cooperative aerial manipulation using multirotors with multi-dof robotic arms,'' \emph{IEEE/ASME Transactions on Mechatronics}, vol.~23, no.~2, pp. 702--713, 2018.

\bibitem{zhang2019robust}
G.~Zhang, Y.~He, B.~Dai, F.~Gu, J.~Han, and G.~Liu, ``Robust control of an aerial manipulator based on a variable inertia parameters model,'' \emph{IEEE Transactions on Industrial Electronics}, vol.~67, no.~11, pp. 9515--9525, 2019.

\bibitem{lee2020aerial}
D.~Lee, H.~Seo, D.~Kim, and H.~J. Kim, ``Aerial manipulation using model predictive control for opening a hinged door,'' in \emph{2020 IEEE International Conference on Robotics and Automation (ICRA)}.\hskip 1em plus 0.5em minus 0.4em\relax IEEE, 2020, pp. 1237--1242.

\bibitem{bicego2020nonlinear}
D.~Bicego, J.~Mazzetto, R.~Carli, M.~Farina, and A.~Franchi, ``Nonlinear model predictive control with enhanced actuator model for multi-rotor aerial vehicles with generic designs,'' \emph{Journal of Intelligent \& Robotic Systems}, vol. 100, no.~3, pp. 1213--1247, 2020.

\bibitem{kim2017robust}
S.~Kim, S.~Choi, H.~Kim, J.~Shin, H.~Shim, and H.~J. Kim, ``Robust control of an equipment-added multirotor using disturbance observer,'' \emph{IEEE Transactions on Control Systems Technology}, vol.~26, no.~4, pp. 1524--1531, 2017.

\bibitem{chen2020robust}
Y.~Chen, W.~Zhan, B.~He, L.~Lin, Z.~Miao, X.~Yuan, and Y.~Wang, ``Robust control for unmanned aerial manipulator under disturbances,'' \emph{IEEE Access}, vol.~8, pp. 129\,869--129\,877, 2020.

\bibitem{lee2022rise}
D.~Lee, J.~Byun, and H.~J. Kim, ``Rise-based trajectory tracking control of an aerial manipulator under uncertainty,'' \emph{IEEE Control Systems Letters}, vol.~6, pp. 3379--3384, 2022.

\bibitem{liang2021low}
J.~Liang, Y.~Chen, N.~Lai, B.~He, Z.~Miao, and Y.~Wang, ``Low-complexity prescribed performance control for unmanned aerial manipulator robot system under model uncertainty and unknown disturbances,'' \emph{IEEE Transactions on Industrial Informatics}, vol.~18, no.~7, pp. 4632--4641, 2021.

\bibitem{liang2024observer}
X.~Liang, Y.~Wang, H.~Yu, Z.~Zhang, J.~Han, and Y.~Fang, ``Observer-based nonlinear control for dual-arm aerial manipulator systems suffering from uncertain center of mass,'' \emph{IEEE Transactions on Automation Science and Engineering}, 2024.

\bibitem{li2024finite}
H.~Li, Z.~Li, F.~Song, X.~Yu, X.~Yang, and J.~J. Rodr{\'\i}guez-Andina, ``Finite-time fast adaptive backstepping attitude control for aerial manipulators based on variable coupling disturbance compensation,'' \emph{IEEE Transactions on Industrial Electronics}, 2024.

\bibitem{chen2022adaptive}
Y.~Chen, J.~Liang, Y.~Wu, Z.~Miao, H.~Zhang, and Y.~Wang, ``Adaptive sliding-mode disturbance observer-based finite-time control for unmanned aerial manipulator with prescribed performance,'' \emph{IEEE Transactions on Cybernetics}, vol.~53, no.~5, pp. 3263--3276, 2023.

\bibitem{kim2016vision}
S.~Kim, H.~Seo, S.~Choi, and H.~J. Kim, ``Vision-guided aerial manipulation using a multirotor with a robotic arm,'' \emph{IEEE/ASME Transactions On Mechatronics}, vol.~21, no.~4, pp. 1912--1923, 2016.

\bibitem{liang2022adaptive}
J.~Liang, Y.~Chen, Y.~Wu, Z.~Miao, H.~Zhang, and Y.~Wang, ``Adaptive prescribed performance control of unmanned aerial manipulator with disturbances,'' \emph{IEEE Transactions on Automation Science and Engineering}, vol.~20, no.~3, pp. 1804--1814, 2023.

\bibitem{orsag2018aerial}
M.~Orsag, C.~Korpela, P.~Oh, S.~Bogdan, and A.~Ollero, \emph{Aerial manipulation}.\hskip 1em plus 0.5em minus 0.4em\relax Springer, 2018.

\bibitem{arleo2013control}
G.~Arleo, F.~Caccavale, G.~Muscio, and F.~Pierri, ``Control of quadrotor aerial vehicles equipped with a robotic arm,'' in \emph{21St Mediterranean Conference on Control and Automation}.\hskip 1em plus 0.5em minus 0.4em\relax IEEE, 2013, pp. 1174--1180.

\bibitem{spong2008robot}
M.~W. Spong and M.~Vidyasagar, \emph{Robot Dynamics and Control}.\hskip 1em plus 0.5em minus 0.4em\relax John Wiley \& Sons, 2008.

\bibitem{mellinger2011minimum}
D.~Mellinger and V.~Kumar, ``Minimum snap trajectory generation and control for quadrotors,'' in \emph{2011 IEEE International Conference on Robotics and Automation (ICRA)}.\hskip 1em plus 0.5em minus 0.4em\relax IEEE, 2011, pp. 2520--2525.

\bibitem{lee2010geometric}
T.~Lee, M.~Leok, and N.~H. McClamroch, ``Geometric tracking control of a quadrotor {UAV} on se (3),'' in \emph{49th IEEE Conference on Decision and Control (CDC)}.\hskip 1em plus 0.5em minus 0.4em\relax IEEE, 2010, pp. 5420--5425.

\bibitem{khalil2002nonlinear}
H.~K. Khalil, \emph{Nonlinear systems}.\hskip 1em plus 0.5em minus 0.4em\relax Prentice hall Upper Saddle River, NJ, 2002, vol.~3.

\end{thebibliography}
\end{document}